\documentclass[]{bytedance_seed}

\usepackage[toc,page,header]{appendix}

\usepackage{natbib}

\usepackage{CJKutf8}

\usepackage{xargs}  

\usepackage{todonotes}  

\usepackage{multirow}

\usepackage{cleveref}

\usepackage{amsmath}
\usepackage{dsfont}

\usepackage{subcaption}

\usepackage{svg}

\usepackage{mathrsfs}
\usepackage{adjustbox}
\usepackage{multirow}
\usepackage{multirow}
\usepackage{multicol}
\usepackage{tcolorbox}
\usepackage{changepage}
\usepackage{graphicx}
\usepackage{amssymb}
\usepackage{array}
\usepackage{bm}
\usepackage{hyperref}

\usepackage{minitoc}

\title{Seedance 1.0: Exploring the Boundaries of \\ Video Generation Models}

\author[]{ByteDance Seed}

\abstract{
Notable breakthroughs in diffusion modeling have propelled rapid improvements in video generation, yet current foundational model still face critical challenges in simultaneously balancing prompt following, motion plausibility, and visual quality.
In this report, we introduce \textbf{Seedance 1.0}, a high-performance and inference-efficient video foundation generation model that integrates several core technical improvements: 
(i) multi-source data curation augmented with precision and meaningful video captioning, enabling comprehensive learning across diverse scenarios; 
(ii) an efficient architecture design with proposed training paradigm, which allows for natively supporting multi-shot generation and jointly learning of both text-to-video and image-to-video tasks.
(iii) carefully-optimized post-training approaches leveraging fine-grained supervised fine-tuning, and video-specific RLHF with multi-dimensional reward mechanisms for comprehensive performance improvements; 
(iv) excellent model acceleration achieving ~10× inference speedup through multi-stage distillation strategies and system-level optimizations. Seedance 1.0 can generate a 5-second video at 1080p resolution only with 41.4 seconds (NVIDIA-L20). Compared to state-of-the-art video generation models, Seedance 1.0 stands out with high-quality and fast video generation having superior spatiotemporal fluidity with structural stability, precise instruction adherence in complex multi-subject contexts, native multi-shot narrative coherence with consistent subject representation.
Seedance 1.0 is now accessible on \href{https://console.volcengine.com/ark/region:ark+cn-beijing/experience/vision?type=GenVideo}{Volcano Engine}\textsuperscript{$\alpha$}.
}

\checkdata[Official Page]{\url{ https://seed.bytedance.com/seedance} }
\checkdata[\textsuperscript{$\alpha$}Model ID]{Doubao-Seedance-1.0-pro}

\begin{document}
\begin{CJK*}{UTF8}{gbsn}

\maketitle

\definecolor{chinese_red}{HTML}{8B4513}
\definecolor{english_blue}{HTML}{4169E1}

\begin{figure}[ph]
\begin{center}
\vspace{-25pt}
\includegraphics[height=6.8cm]{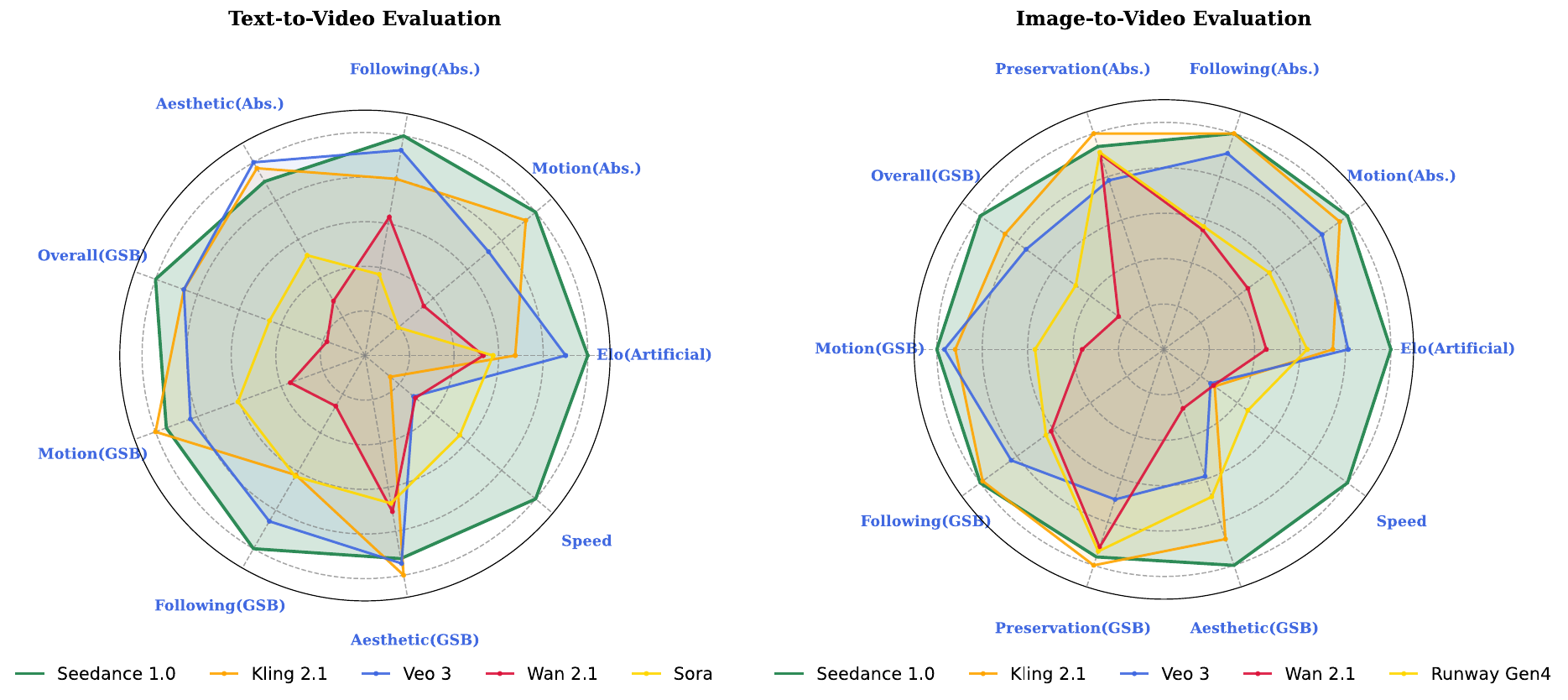}
\end{center}
\label{fig:overall_eval}
\vspace{-10pt}
\caption{Overall evaluation. Left: Text-to-Video; Right: Image-to-Video. Seedance 1.0 ranks first on both the two video generation leaderboards of Artificial Analysis on Jun 10, 2025 (Due to unavailable public data, the Elo score for Kling 2.1 is taken from Kling 2.0). "Speed" denotes the inverse of the average generation time per second of video (from API).}
\vspace{-8pt}
\end{figure}%



\clearpage

\tableofcontents

\newpage

\section{Introduction}

With recent advances in diffusion models, the progress of video generation has been accelerated considerably. Leading open-source frameworks including Wan~\citep{wan2025wan}, Huanyuan Video~\citep{kong2024hunyuanvideo}, and CogVideoX~\citep{yang2024cogvideox}, complemented by commercial systems such as Veo and Keling, have catalyzed broad academic and industrial adoption. However, current video generation foundation models still have critical challenges in balancing multidimensional requirements, particularly in prompt following, motion plausibility, and visual fidelity.
To address these limitations, we present \textit{\textbf{Seedance 1.0}}, a foundational video generation model with native support bilingual (Chinese/English) video generation and multi-task versatility encompassing text-to-video synthesis and image-guided video generation. Seedance 1.0 integrates four key technical improvements:

\begin{itemize}[leftmargin=*]

 \item \textbf{\textit{Multi-Source Data with Comprehensive Video Captioning.}} Through multi-stage, multi-perspective curation and dataset balancing, we construct a large-scale high-quality video dataset spanning diverse categories, styles, and sources. This enables a comprehensive learning of rich scenarios, topics, and action dynamics. Our precision video captioning system ensures accurate interpretation of user instructions while enabling fluent generation of complex video narratives.

 \item \textbf{\textit{Efficient Architecture Design.}}  
 In our design, we decouple spatial and temporal layers with an interleaved multimodal positional encoding. This allows our model to jointly learn both text-to-video and image-to-video in a single model, and natively support multi-shot video generation. In particular, the decoupled layers are integrated with carefully-designed window attentions which further improve model efficiency considerably in both training and inference.


 \item \textbf{\textit{Enhanced Post-Training Optimization.}}  We use a small set of carefully collected data for SFT, which is followed by a video-tailored RLHF algorithm (Reinforcement Learning from Human Feedback). We develop feedback-driven learning algorithms using multiple well-developed reward models, which allow us to considerably improve our performance on both T2V and I2V, in terms of motion naturalness, structural coherence, and visual fidelity.


 \item \textbf{\textit{Inference Acceleration.}} We proposed a multi-stage distillation framework to reduce the number of function evaluations (NFE) required for generation, with inference infrastructure optimization techniques, achieving over 10× end-to-end speedup with no degradation in model performance. 

\end{itemize} 

Compared with contemporary models, Seedance 1.0 exhibits four distinguishing characteristics:

\begin{itemize}[leftmargin=*]

\item \textbf{\textit{Comprehensive Generation Capabilities.}}
Seedance 1.0 achieves superior spatiotemporal coherence and structural stability, demonstrating exceptional motion fluidity and physical plausibility. The model produces photorealistic visuals with nuanced textures and compositional richness, attaining state-of-the-art performance across both proprietary evaluation suites and authoritative third-party benchmarks.

\item \textbf{\textit{Precision Instruction Following.}}
Through comprehensive learning of diverse scenarios, entities, and action semantics, Seedance 1.0 precisely interprets complex user specifications. It robustly handles multi-agent interactions, adaptive camera control, and stylistic variations while maintaining narrative continuity.

\item \textbf{\textit{Multi-Shot Narrative Capability.}} Seedance 1.0 natively supports coherent multi-shot storytelling with stable view transitions while maintaining consistent subject representation across temporal-spatial transformations.

\item \textbf{\textit{Ultra-Fast Generation Experience.}} With multiple model acceleration techniques, Seedance 1.0 significantly reduces inference costs: it can generate a 5-second video at 1080p resolution only with 41.4 seconds (NVIDIA-L20), which is substantially faster than other commercial counterparts.

\end{itemize} 
Seedance 1.0 will be integrated into multiple platforms in June 2025, including Doubao\footnotemark[1] and Jimeng\footnotemark[2]. We envision it becoming an essential productivity tool for enhancing professional workflows and daily creative applications.

\footnotetext[1]{https://www.doubao.com/chat/create-video}
\footnotetext[2]{https://jimeng.jianying.com/ai-tool/video/generate}

\section{Model Design}

\subsection{Variational Autoencoder}
Variational autoencoders (VAEs)~\citep{Kingma2013AutoEncodingVB} are widely adopted in modern large-scale image and video generation models~\citep{rombach2022high} to reduce the computation of the subsequent diffusion model and facilitate efficient training and inference. Typically, a variational auto-encoder is usually composed of an encoder and a decoder; the encoder compresses the raw redundant pixel information into a compact latent representation, while the decoder reconstructs the original input from these latent features. The quality of VAE reconstruction directly establishes an upper bound for the realism and clarity achievable by the generative process, whereas the distribution of latent representations significantly impacts the convergence behavior of subsequent Diffusion Transformers (DiT). 

\textbf{Temporally-Causal Compression.} Following MAGVIT \cite{yu2023language}, we adopt a temporally causal convolutional architecture for both the encoder and decoder, allowing joint spatial-temporal compression of images and videos within latent space. To be more specific, the model transforms the input data from the RGB pixel space with shape $(T'+1, H', W', 3)$ into a continuous latent representation with shape $(T+1, H, W, C)$, where $(t, h, w, c)$ denotes time, height, width and channel dimensions with $r_t = \frac{T'}{T}$, $r_h = \frac{H'}{H}$, and $r_w = \frac{W'}{W}$ representing the downsample ratios along these three axes, respectively. Benefiting from the causal design, the VAE model can seamlessly process image input and output in the case of $T=T'=0$. The overall compression ratio is given by
\begin{equation}
    r = \frac{C \times T \times H \times W}{3 \times T' \times H' \times W'} = \frac{C}{3 \times r_t \times r_h \times r_w}.
    \label{equa:compression_ratio}
\end{equation}
In our practice, for the sake of training and inference efficiency and overall reconstruction and generation performance, we set $(r_t, r_h, r_w) = (4,16,16)$ and $C=48$. To accommodate the higher downsampling rate and pursue better generation performance, we remove the patchification operation on the DiT side, following the strategy adopted in DCAE \cite{chen2024deep}.

\textbf{VAE Training.} Our VAE is trained with $L1$ reconstruction loss, KL loss, LPIPS \cite{lpips_zhang2018unreasonable} perceptual loss and adversarial training loss.  Adversarial training has shown to be effective in improving the quality of VAE reconstruction by enforcing finer supervision on local textures and detailed structures. Taking into account appearance and motion modeling simultaneously, we apply a hybrid discriminator with an architecture similar to that used in PatchGAN \cite{isola2018imagetoimagetranslationconditionaladversarial}.

\subsection{Diffusion Transformer}
With the visual tokens encoded by VAE and text tokens generated by a text encoder, we employ the transformer as our diffusion backbone~\cite{peebles2023scalable}, where a fine-tuned decoder-only LLM as the text encoder. The visual tokens are then concatenated with textual tokens and fed into the transformer blocks.

\textbf{Decoupled Spatial and Temporal Layers.} Considering both training and inference efficiency, we build the diffusion transformer with decoupled spatial and temporal layers, where the spatial layers perform attention aggregation within each frame, while the temporal layers focus attention computation across frames. We perform window partition within each frame in the temporal layers, allowing for a global receptive field across the temporal dimension. In addition, textual tokens only participate in cross-modality interaction in spatial layers.

\begin{figure*}[t]
\centering
\includegraphics[width=\textwidth]{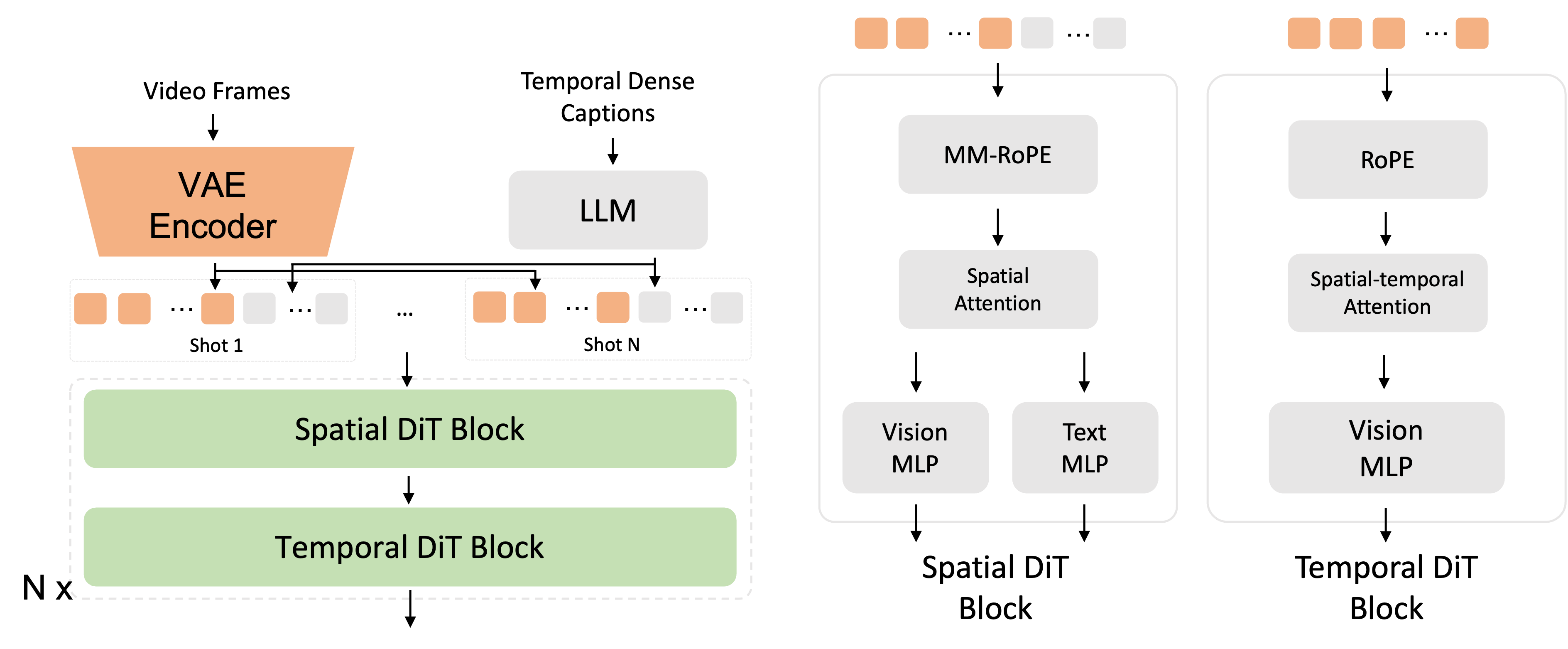}
\caption{Our diffusion transformer architecture.}
\end{figure*}

\textbf{MMDiT Architecture.} For the transformer blocks, we follow the MMDiT design in Stable Diffusion 3~\cite{esser2024scaling}, where a multi-modality self-attention layer is applied exclusively in spatial layers to integrate both the visual and textual tokens, whereas a self-attention layer only processes the visual tokens in temporal layers. Considering the semantic differences between visual and textual tokens, we use two separate sets of weights including adaptive layer norm, QKV projection, and MLP, for the two modalities in spatial layers. To prevent training instability, the Q and K embeddings are normalized prior to the attention matrix calculation.

\textbf{Multishot MM-RoPE.} In this paper, in addition to using 3D RoPE encoding for visual tokens, following Seaweed~\cite{seawead2025seaweed} and LCT~\cite{guo2025long}, we add 3D Multi-modal RoPE~(MM-RoPE) in the concatenated sequences by adding extra 1D positional encoding for textual tokens. The MM-RoPE also supports interleaved sequences of visual tokens and textual tokens, and can be extended to training video with multiple shots, where shots are organized in the temporal order of actions and each shot has its own detailed caption.

\textbf{Unified Task Formulation.} To enable conditional video generation, we concatenate the noisy inputs with cleaned or zero-padded frames along the channel dimension, and use binary masks to indicate which frames are instructions to follow~\cite{girdhar2024factorizing}. With this formulation, we can further unify different generation tasks such as text-to-image, text-to-video and image-to-video \cite{chen2023control}. During the training process, we mix these tasks and adjust the proportion by controlling the conditional inputs.

\subsection{Diffusion Refiner}
Take into account the training and inference efficiency, we employ a cascaded diffusion framework for high-resolution (HR) video generation. The base model generates 480p videos first,  which are then upscaled to 720p or 1080p high-resolution videos through a learned diffusion refiner model to enhance visual details and textures. 

\textbf{Refiner Model Training.} To facilitate training, the diffusion refiner model is initialized from the pre-trained base model. Different from the base model, the diffusion refiner model is trained with conditioning on the low-resolution 
(LR) videos. Specifically, the LR video is upsampled to a high resolution first, then concatenated with the diffusion noise along the channel dimension to form the input of the diffusion transformer.

\subsection{Prompt Engineering (PE)}
\label{sec:pe}
As described in Sec~\ref{sec:caption}, texts used in DiT are form of dense video captions. Therefore, we need to employ a large language model to convert the user prompts into corresponding caption format. To achieve this, we initialize based on Qwen2.5-14B~\cite{yang2024qwen2_5} and employ two stages to implement high-quality Prompt Engineering (PE): Supervised Fine-Tuning (SFT) and Reinforcement Learning (RL).

\textbf{Supervised Fine-Tuning.} In the SFT stage, we synthesize large amount of user prompts and their dense caption expression by manual annotation. We specially devide the image-to-video (i2v) and text-to-video (t2v) tasks, as they are different in user prompt styles. We then adpot a fully fine-tuning strategy to train the model on the annotated data to aquire basic rephrasing abilitity.

\textbf{Reinforcement Learning.} However, due to the presence of model hallucinations, the results of the first SFT stage cannot guarantee that the semantics of the rewritten results fully meet the requirements of the user prompts. Therefore, we carefully collect a dataset of pairs with correct and incorrect rephrased results to perform the Direct Preference Optimization (DPO)~\cite{rafailov2023dpo,ji2024prompt} training. In this stage, we used the Low-Rank Adaptation (LoRA)~\cite{hu2022lora} fine-tuning strategy on the SFT model.

After the above stages, our prompt engineering model has strong ability to understand user prompts and gives precise and high-quality rephrased results in video caption format, consistent with DiT training.

\section{Data}
\label{sec:data}

\begin{figure*}[t]
\centering
\includegraphics[width=\textwidth]{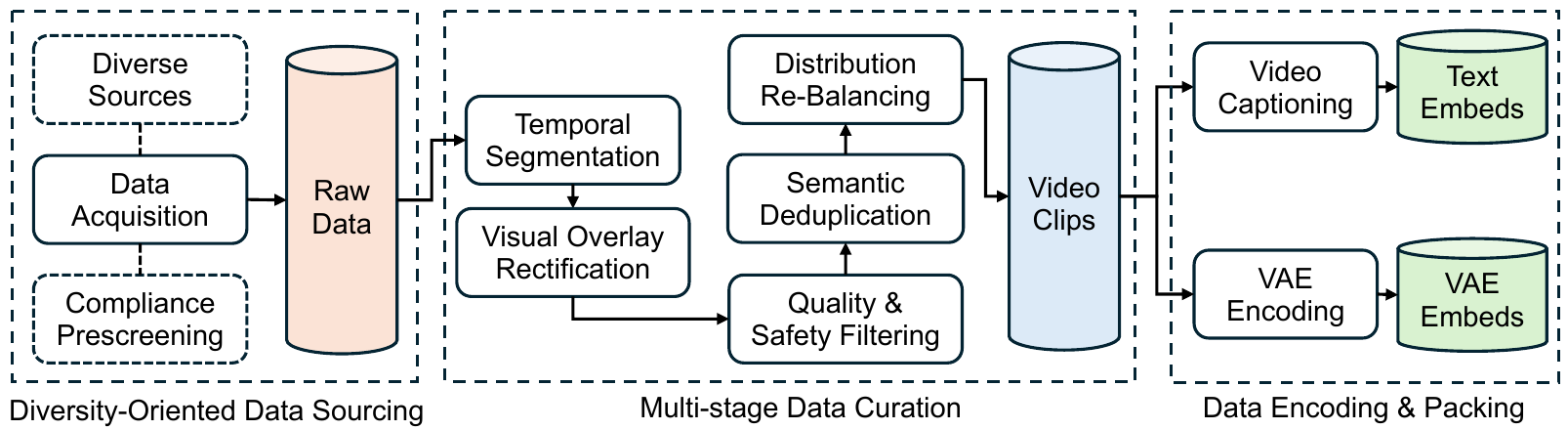}
\caption{
Our video data processing pipeline, transforming heterogeneous raw videos into a refined, feature-rich training dataset. The workflow comprises three main phases:
(1) Diversity-oriented data sourcing for initial acquisition and compliance prescreening from diverse sources;
(2) Multi-stage data curation refines raw data into video clips; and
(3) Offline data packing where video captioning and VAE encoding are used to generate text and VAE embeddings for model training.}
\label{fig:data_processing}
\end{figure*}

The performance of video generation models is inextricably linked to the scale, diversity, and quality of the training data. While our broader training corpus incorporates both video and image datasets, with image data preparation following methodologies similar to Seedream~\cite{gong2025seedream}, this section specifically details our rigorous approach to curating video data. We develop a systematic data processing workflow, illustrated in~\cref{fig:data_processing}, to transform vast, heterogeneous raw video collections into a refined, high-quality, diverse, and safe dataset for training robust video generation models. This workflow is deployed as a robust, automated system optimized for high-throughput processing of massive data volumes.

\subsection{Data Pre-Processing}
\label{sec:data_process}

At the heart of our video data curation is a multi-stage pre-processing pipeline, designed to tackle the challenges of raw video collections. Each subsequent stage systematically elevates the dataset's standard, preparing it for robust model training. The following paragraphs detail each component of this comprehensive pipeline, which ensures that only video clips meeting our stringent criteria contribute to the final dataset.

\textbf{\textit{Diversity-Oriented Data Sourcing.}}  
Our video data acquisition strategy prioritizes ethically and legally sourced content from diverse public and licensed repositories. We aim to maximize coverage across critical dimensions, including clip duration, resolution, subject matter (e.g., humans, animals, objects), scene types (e.g., natural landscapes, urban environments), subject actions, genres (e.g., documentary, animation), artistic styles, camera kinematics, and cinematographic techniques. Raw video collections exhibit significant heterogeneity and often contain undesirable elements, posing key challenges that our pipeline is designed to address.

\textbf{\textit{Shot-Aware Temporal Segmentation.}}  
Raw long-form videos are not suitable for direct model training. We employ automated shot boundary detection techniques by analyzing inter-frame visual dissimilarities or utilizing pre-trained detectors to identify natural scene transitions. Subsequently, videos are segmented into shorter clips, with a maximum duration of $12$ seconds. Each resulting clip may contain one or multiple temporally coherent shots, preserving local narrative flow while ensuring manageable input lengths for model ingestion.

\textbf{\textit{Visual Overlay Rectification.}}  
Many source videos contain extraneous visual overlays such as logos, watermarks, subtitles, or on-screen graphics that can introduce noise or bias. Our rectification stage identifies these occlusions using a hybrid approach of heuristic rule-based systems and specialized object detection models. Frames are then adaptively cropped to maximize the retention of the primary visual content, yielding cleaner and more focused video data.

\textbf{\textit{Quality and Safety Filtering.}}  
To ensure the model is trained on high-quality and ethically compliant data, we enforce rigorous filtering via visual assessment and safety screening. First, clips exhibiting visual defects such as blurriness, excessive jittering, low aesthetic quality, poor cinematographic composition, or predominantly static content are systematically identified and removed by our specialized visual quality model. 
Second, we rigorously exclude harmful or inappropriate material, deploying advanced classifiers to detect content pertaining to pornography, explicit violence, child exploitation, and explicit nudity, thereby ensuring ethical compliance and dataset safety.

\textbf{\textit{Semantic Deduplication.}}  
To promote dataset diversity and prevent model overfitting to redundant content, we perform semantic deduplication. Video clips are represented by robust feature embeddings extracted from an internally developed video representation model, and these embeddings enable clustering of visually and semantically similar clips. Within each identified cluster of near-duplicates, only the single instance with the highest overall quality score (from the preceding quality filtering stage) is retained.

\textbf{\textit{Distribution Rebalancing.}}  
Raw data often exhibits significant category imbalance across various attributes. We analyze the dataset’s distribution along these dimensions by quantifying frequencies across attributes tailored to different semantic and technical perspectives, such as subject categories, scene types, dominant actions, genres, visual styles, clip duration, resolution, and motion characteristics. For over-represented head categories, downsampling is applied. Conversely, for under-represented tail categories, we increase their sampling probability during training and initiate targeted data acquisition to augment their presence, aiming for a more equitable and comprehensive representation of the visual world.

\subsection{Video Captioning}
\label{sec:caption}

Video captions largely affect the instruction-following capabilities of the video generation model. We mainly improve the quality and accuracy of captions to ensure that important content and actions can be seen and described proprerly.

\textbf{Caption Style.} We adopt a dense caption style integrating dynamic and static features. For dynamic features, we meticulously describe actions and camera movements of a video clip, highlighting changing elements. For static features, we elaborate on the characteristics of core characters or scenes in the video.

\textbf{Caption Elements.} We define specific categories dynamic and static features respectively. dynamic features cover categories of motions, subjects or scenes changing and camera movements, while static features include appearances, aesthetics, styles, etc. We collect diverse data on such categories and conduct high-quality manual annotations for training. The trained caption model can accurately describe the critical content of complex and abstract video materials.

\textbf{Model Training.} We train the caption model on the annotated data with Tarsier2~\cite{yuan2025tarsier2advancinglargevisionlanguage}, a model with strong video understanding capabilities. The visual encoder is frozen and the language model is fully fine-tuned. We train on both Chinese and English data to acquire bilingual capabilities. 

During inference, we use our PE model described in Sec~\ref{sec:pe} to rephrase user prompts into detail video captions, in which the format is aligned with the training captions in content and structure.

\subsection{Efficient Engineering Infrastructure}
\label{sec:engineering_infra}

\begin{figure*}[t]
\centering
\includegraphics[width=0.9\textwidth]{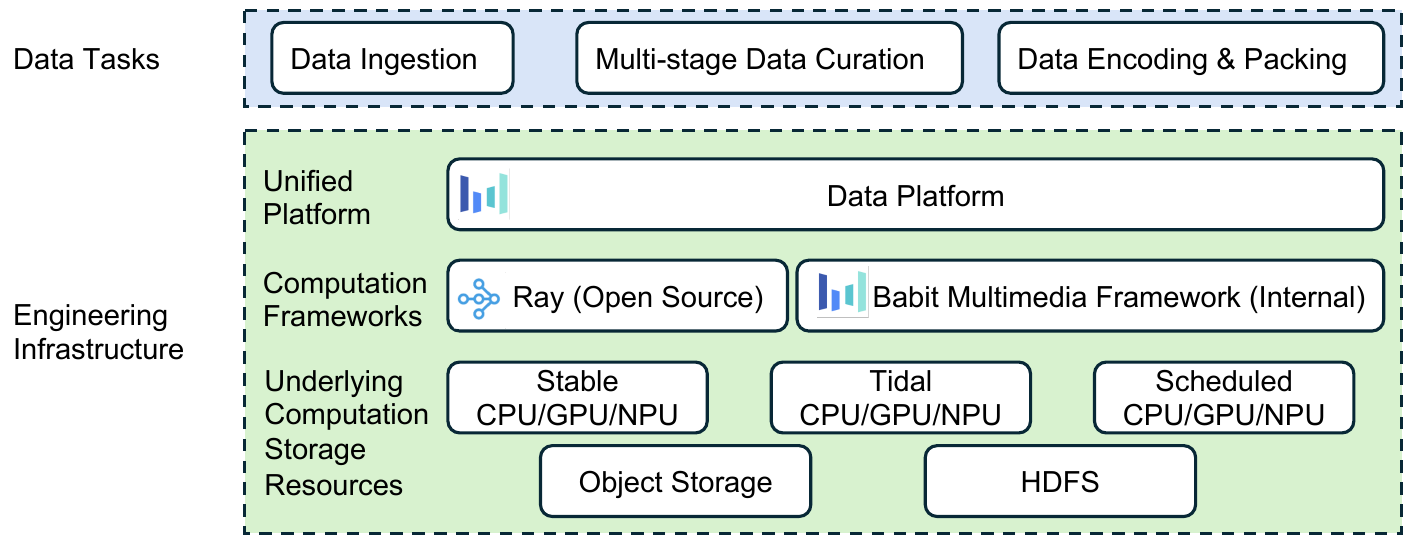}
\caption{
Overview of our engineering infrastructure for data processing.
}
\label{fig:data_engineering}
\end{figure*}

\textbf{\textit{Engineering Infrastructure Overview.}}
Our engineering infrastructure for data processing is illustrated in~\cref{fig:data_engineering},  which consists of three layers: 
at the top is the unified platform layer, automating human-in-the-loop workflows, managing tasks, visualizing data, and monitoring pipelines, etc.; 
in the middle is the computation framework layer, which employs BMF~\cite{bmf} and Ray~\cite{ray} for heterogeneous computing across CPU/GPU/NPU architectures and optimizes resource allocation for both stable and elastic computing; 
at the bottom is the underlying resources layer, which leverages cloud infrastructure from ByteCloud (internal) and Volcengine (external).

\textbf{\textit{Efficient Heterogeneous Computing.}}
\label{sec:engineering_infra_computing}
To maximize resource utilization, our frameworks dynamically allocate video operations to optimal hardware (e.g., CPU for decoding, GPU for deep model inference). Asynchronous communication between computation units is used to mitigate bottlenecks introduced by the performance gap between different types of computation hardware.
To address the complexities arising from the instability of elastic computation resources, our frameworks incorporate two critical capabilities: adaptive auto-scaling to handle resource fluctuations and failure retry mechanisms for preempted tasks. 
Customized versions of BMF and Ray implement these optimizations, delivering near-linear scalability and extremely high throughput to efficiently process massive-scale video training data.

\section{Model Training}

\begin{figure*}[hb]
\centering
\includegraphics[width=\textwidth]{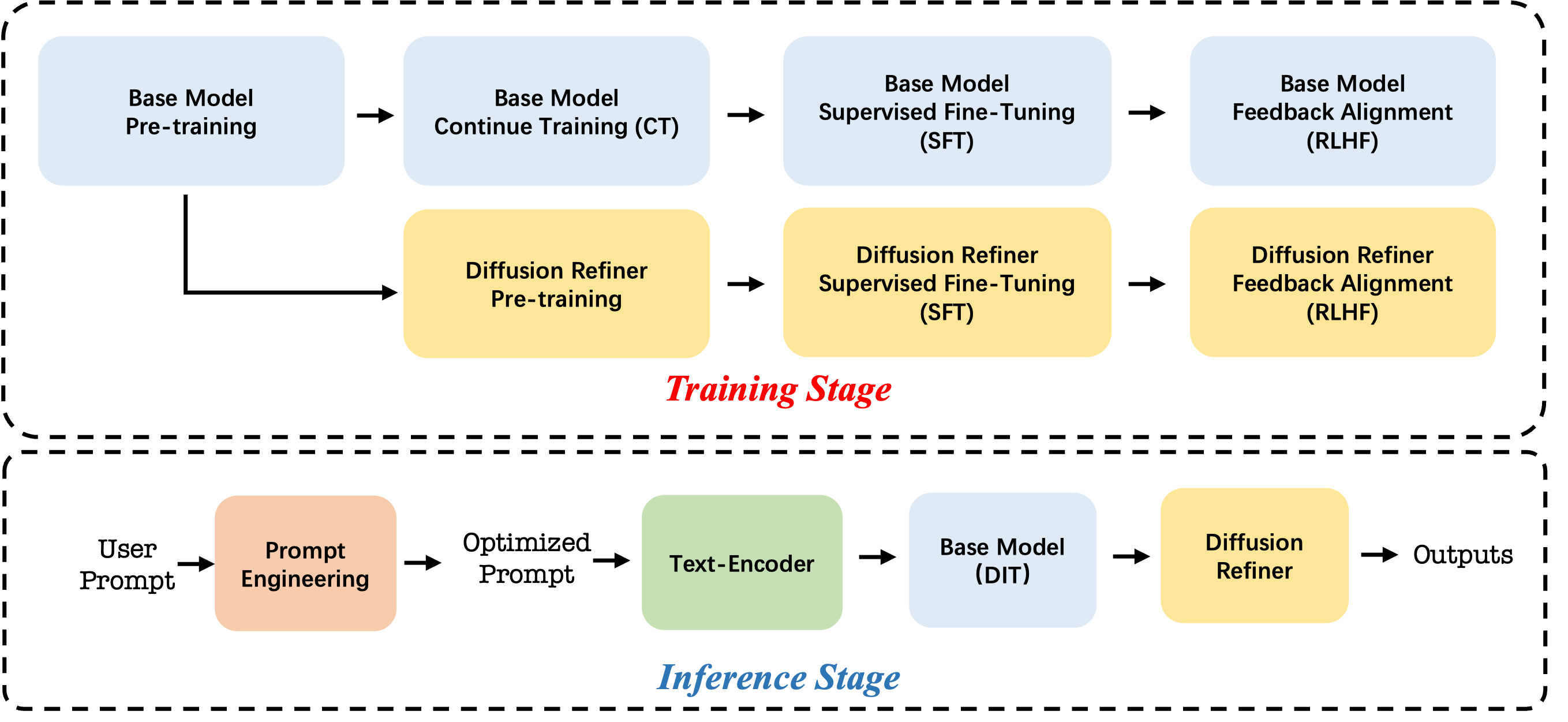}
\caption{Overview of training and inference pipeline.}
\label{fig:stage}
\end{figure*}

\begin{figure*}[hp]
\centering
\includegraphics[width=1.0\linewidth]{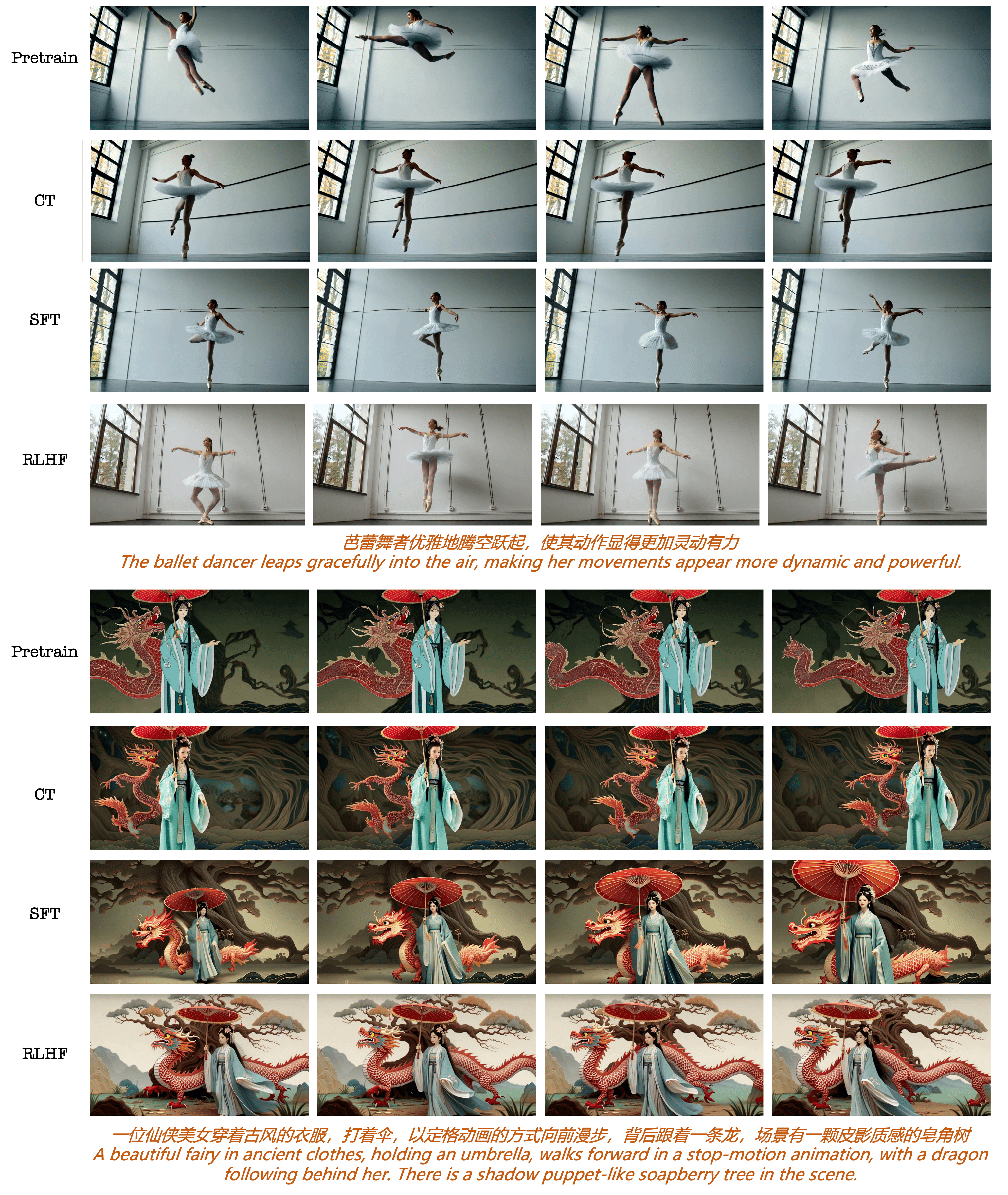}
\vspace{-5pt}
\caption{Visualization during different post-training stages.}
\label{fig:stage_compare}
\vspace{-5pt}
\end{figure*}

As shown in Figure \ref{fig:stage}, we present our training and inference stages of Seedance 1.0. Our training process is divided into several substages, including pre-training, continue training (CT), supervised fine-tuning (SFT) and human feedback alignment (RLHF). Our refiner also includes pre-training, SFT and RLHF. The visualization results during different training stages are presented in Figure \ref{fig:stage_compare}, where each stage can progressively improve the results.

\subsection{Pre-Training}
\textbf{Diffusion Scheduling.} During training, we employ the flow matching framework with velocity prediction, and a training timestep is sampled from a logit-normal distribution. Considering that videos with higher resolution and longer duration require more noise to disrupt their signal, we then transform the training timestep with a resolution-aware shift, which increases the noise perturbation for videos with higher resolution and longer duration.

\textbf{Progressive Training.} To enable higher data throughput and training efficiency, we initialize the model with sufficient low-resolution text-to-image~($256 px$) training and then progressively introduce video modalities with higher resolution and higher fps in following stages: (1) We conduct image-video joint training using $256 px$ images and video clips from 3 to 12 seconds~($12$ fps). (2) In the second stage, we increase the training resolution to $640 px$ while maintaining the same duration. (3) In the final stage, we train the models with $24$ fps video to further improve the video smoothness. During video pre-training, we also retain a small portion of text-to-image task to maintain semantic alignment and set the proportion of the image-to-video task to 20\% to activate the ability to follow visual prompts.

\subsection{Continue Training (CT)}
As the image-to-video task constitutes only a small fraction of pre-training, the model's potential in this area remains underexplored. To address this, we introduce the Continue Training (CT) phase focused on strengthening image-to-video generation after pre-training. In this phase, we increase the image-to-video ratio from $20\%$ to $40\%$ and further refine the training dataset to improve overall multitask performance.

\textbf{High-Quality Data Selection.} We select a subset of the pre-training data with higher aesthetic quality and richer motion dynamics by using a series of specialized evaluation models, including aesthetic scorer and motion evaluators based on optical flow. Since the first frame is always provided in the image-to-video task, we design two types of caption for training: (1) original long captions with detailed descriptions of both dynamic and static content, and (2) short captions that focus solely on motion dynamics by removing the static description corresponding to the first frame. This encourages stronger semantic alignment with the training objective.

\textbf{Training Strategy.} During continued training, we use slightly fewer GPUs than in the pre-training stage, while maintaining an annealed learning rate schedule. The richer motion dynamics and diverse captions enable the model to generate more natural and smoother videos. Furthermore, the higher aesthetic quality of the training data leads to significant improvements in the visual fidelity of text-to-video generation. As a result, the final model supports both text-to-video and image-to-video tasks with enhanced overall performance.

\subsection{Supervised Fine-Tuning (SFT)}
Following CT, we perform supervised fine-tuning (SFT) to further align the model's output with human preferences regarding visual quality and motion coherence. During this phase, the model trains on a carefully curated set of high-quality video-text pairs with manually verified captions, allowing it to generate videos with improved aesthetics and more consistent motion dynamics.

\textbf{Human-Curated Dataset.} Ensuring data quality and distributional balance is essential. To achieve this, we define several hundred categories based on visual style, motion type, and other key attributes. We then collect data in a targeted manner within each category, resulting in a curated dataset of high-quality video samples with accurate and meaningful captions.

\textbf{Model Merging.} To fully leverage high-quality data, we train separate models on curated subsets designed to capture a wide range of styles, motions, and scenarios. The resulting models are subsequently merged into a single model that integrates their respective strengths. Each model is trained with a smaller learning rate than in pre-training and utilizes a limited number of GPUs. Moreover, we apply early stopping at an effective point to prevent overfitting and maintain text controllability. The final merging step significantly improves both visual fidelity and motion quality.

\subsection{Human Feedback Alignment (RLHF)}

\subsubsection{Feedback Data Infrastructure}

We collect prompts from training datasets and online users, and perform data balancing and information filtering on prompts to discard duplicate and ambiguous ones. We collect high-quality video data pairs for human preference labeling, including synthetic videos generated by different stages of our model.  Experimental results demonstrate that the incorporation of multiple source visual materials can further enhance the domain capacity of the RM model, expand the preference upper bound of RM, and strengthen generalization capabilities. We adopt a multi-dimensional annotation approach in the labeling process, i.e., selecting the best and worst videos under a specific labeling dimension while ensuring that the best videos are not inferior to the worst ones in other dimensions.

\begin{figure*}[t]
\centering
\includegraphics[width=\linewidth]{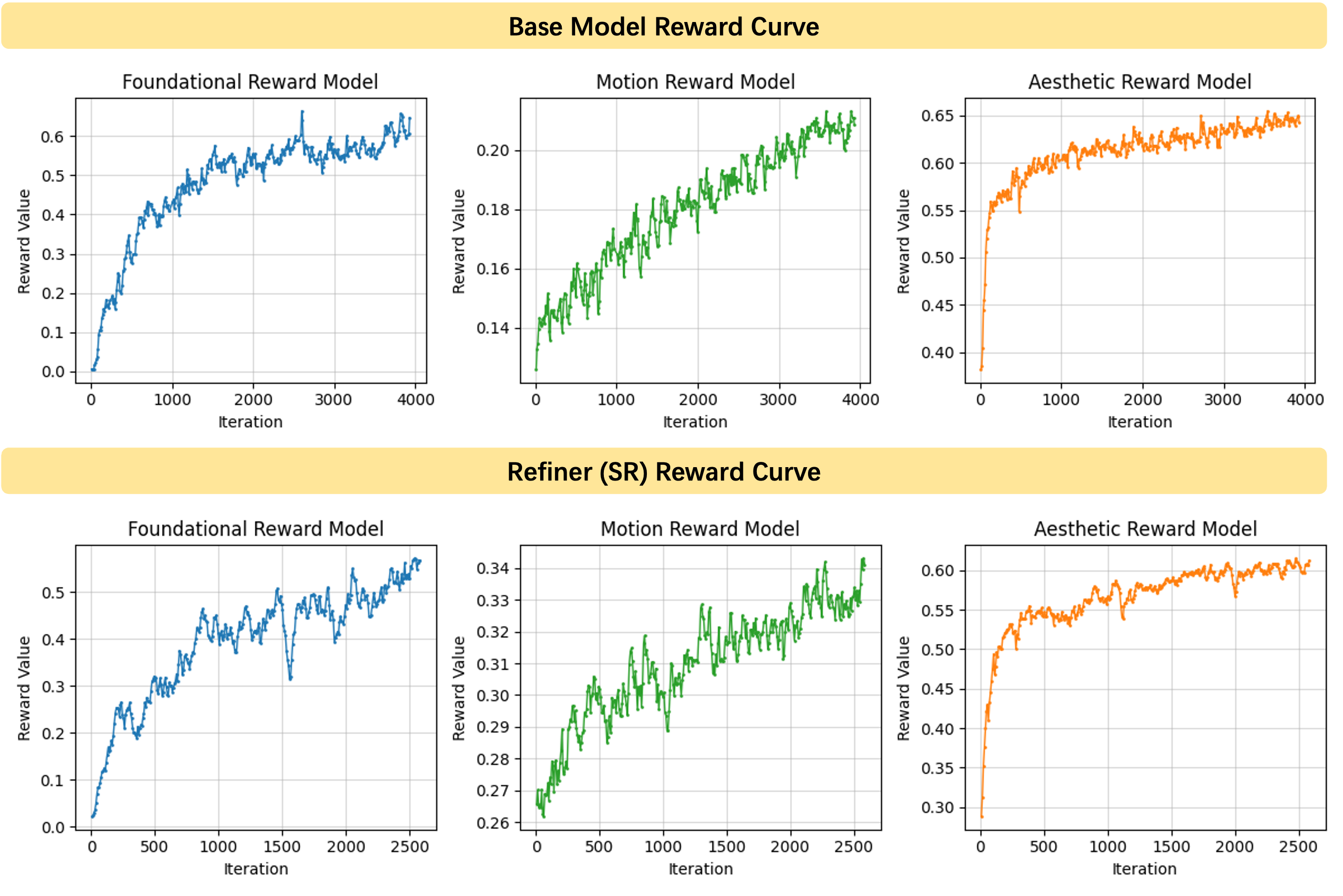}

\caption{The reward curves show that the values across diverse reward models all exhibit a stable and consistent upward trend during the base model and Refiner RLHF process.}
\label{fig:Reward}
\end{figure*}

\subsubsection{Reward Model}

To comprehensively enhance model performance, we design a sophisticated reward system comprising three specialized reward models: Foundational Reward Model, Motion Reward Model, and Aesthetic Reward Model. These dimension-specific reward models, coupled with video-tailored RLHF optimization strategies, enable comprehensive improvements in multiple aspects of the model capabilities, as illustrated in Figure \ref{fig:Reward}. Foundational reward model focuses on enhancing fundamental model capabilities, such as image-text alignment and structural stability. We employ a Vision-Language Model as the architecture of this reward model. Motion reward model helps to mitigate video artifacts while enhancing motion amplitude and vividness.
Given that video aesthetics primarily derive from keyframes, we design the aesthetic reward model from image-space input inspired by Seedream \cite{gao2025seedream,gong2025seedream}, with the data source modified to use keyframes from videos.

\subsubsection{Base Model Feedback Learning}
Reward feedback learning  \cite{xue2025dancegrpo,zhang2024onlinevpo,liu2025flow,liu2025improving} have been widely used in currnet diffusion models. In Seedance 1.0, we simulate the video inference pipeline during training, directly predict $x_0$ (generated clean video) when the Reward Model (RM) adequately assesses video quality. The optimization strategy directly maximizes the composite rewards from multiple RMs. Comparative experiments against DPO/PPO/GRPO demonstrate that our reward maximization approach is the most efficient and effective approach, comprehensively improving text-video alignment, motion quality, and aesthetics. Furthermore, we preform multi-round iterative learning between the diffusion model and RMs. This approach raises the performance bound of the RLHF process and is more stable and controllable than dynamic update of the RM.

\subsubsection{Super-Resolution RLHF Framework}
\begin{figure*}[h]
\centering
\includegraphics[width=0.95\linewidth]{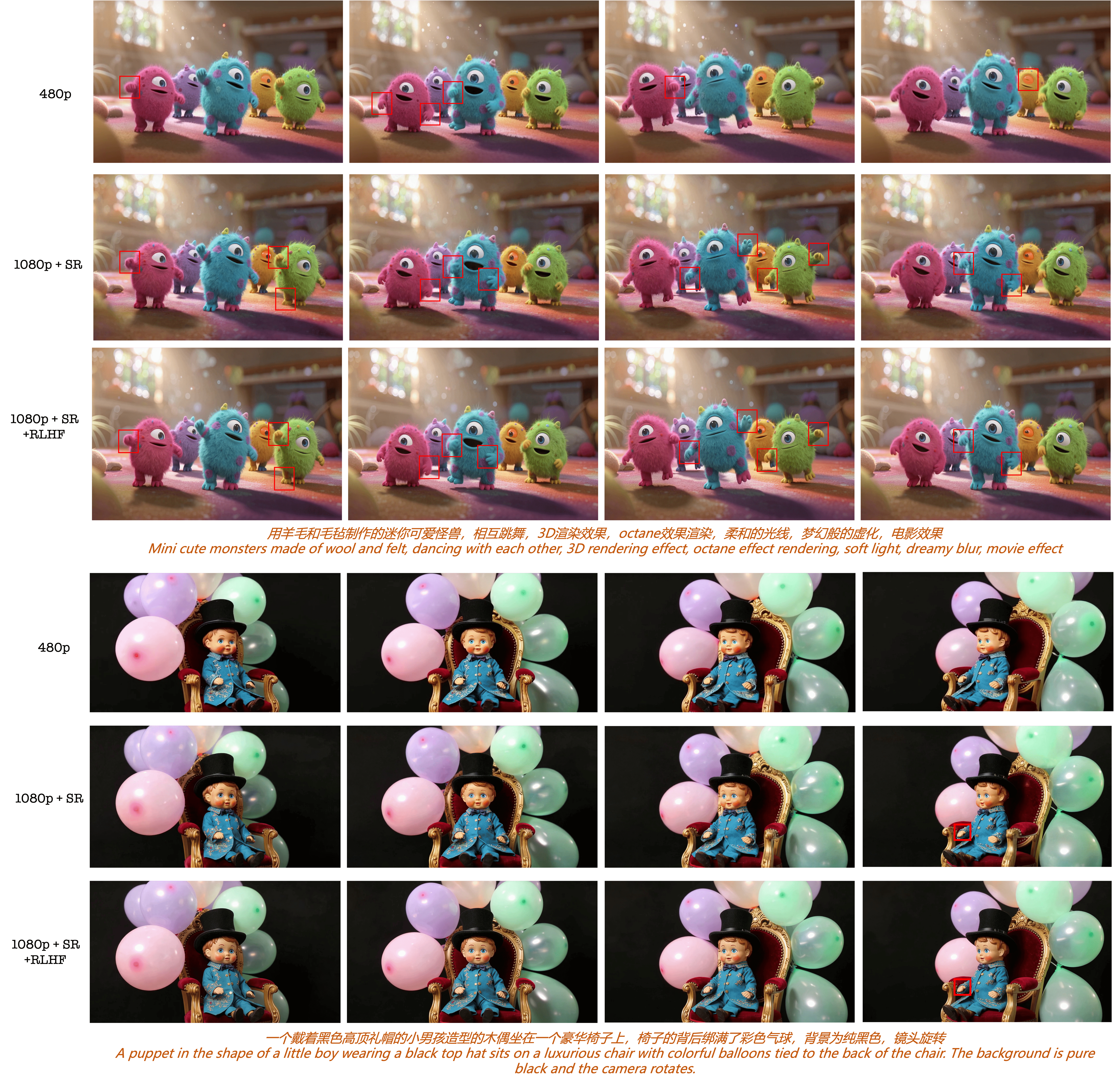}
\vspace{-5pt}
\caption{Visualization during different resolutions and RLHF process.}
\label{fig:sr_compare}
\vspace{-5pt}
\end{figure*}

As shown in Figure \ref{fig:sr_compare}, we also apply RLHF on our diffusion refiner, which can be regarded as a diffusion-based conditional generative model. During training, low-resolution VAE latent space representations serve as conditional inputs to the super-resolution model, while the generated high-resolution videos are evaluated by multiple Reward Models. We directly maximize a linear combination of these reward signals. Notably, our approach applies RLHF directly to the accelerated refiner model, effectively enhancing motion quality and visual fidelity in low-NFE scenarios while maintaining computational efficiency.


\section{Inference Optimizations}

\subsection{Model Acceleration}
\textbf{DiT Optimizations.} To accelerate DiT inference, we adopt diffusion distillation techniques to reduce the number of function evaluations (NFE) required for generation. We incorporate the Trajectory Segmented Consistency Distillation (TSCD) technique, originally introduced in HyperSD\cite{ren2025hyper}, which partitions the denoising trajectory into multiple segments and enforces consistency between predicted and target states across these segments. This allows the student model to learn an accurate approximation of the diffusion process with fewer steps. Using TSCD, our DiT model performs competitively with 4x acceleration, offering a strong balance between speed and fidelity. To push acceleration further, we incorporate Score Distillation from  RayFlow\cite{shao2025rayflow}, which aligns the student model’s predicted noise (i.e., score function) with that of the teacher using expected noise consistency. This approach supports trajectory-level optimization for each sample, enabling more stable and adaptive sampling even at low NFEs. It effectively improves generalization and reduces artifacts during fast generation. To improve visual quality, we extend the adversarial training strategy from APT\cite{lin2025diffusion} to a multi-step distillation setting, incorporating human preference data for supervision. A learned discriminator guides the student model toward outputs favored by human judgments, effectively mitigating artifacts from aggressive acceleration and enhancing perceptual realism.

Through the proposed distillation pipeline, our final distilled model achieves comparable results to the original model across four expert-evaluated dimensions: prompt alignment, motion quality, visual fidelity, and consistency with the source image.

\textbf{VAE Optimizations.} In video generation tasks, the decoding process from latent space to pixel space incurs significant computational cost. We profiled the VAE decoder and found that stages closer to the pixel space dominate the latency. By narrowing the channel widths in these stages, we design a thin VAE decoder. Retraining it with a fixed pre-trained encoder, we achieve a 2× speedup with no loss in visual quality of the end-to-end video generation.

\subsection{Inference Infrastructure}

\textbf{High-Performance Kernel.} Extensive kernel fusion efforts have been conducted on the model's core modules, resulting in a cumulative 15\% improvement in the model's inference throughput.

\textbf{Quantization and Sparse.} Building on the Seedream \cite{gong2025seedream} technical solution, we have implemented fine-grained mixed-precision quantization tailored for Attention and Gemm operations. Moreover, our exploration revealed that the sparse attributes of DiTs exhibit hierarchical and blockified structures across and within various modalities. Expanding on the methodology established by AdaSpa \cite{adaspa}, we have introduced a streamlined tuning solution focused on minimizing search stage overhead. Additionally, we have successfully integrated our optimized fine-grained Attention Quantization approach into this scheme. Numerous efforts have been dedicated to mitigating the impact of full quantization and sparsity on the quality of pixel-level generation. We have achieved an optimal balance between performance and efficiency.

\textbf{Parallelism Strategy.} In order to decrease the allocated massive memory due to the long sequence in video generation schema. A customized adaptive hybrid parallel strategy has been proposed to effectively split the sequences. This approach integrates the concept of context parallelism to optimize communication processes, resulting in a reduction of communication overhead to a quarter of the level observed in Ulysses \cite{jacobs2023deepspeed}. Simultaneously, we have further reduced end-to-end communication overhead by introducing FP8 communication. 

\textbf{Async Offloading Strategy.} Due to the extensive computational demands of attention coupled with the large model size. We developed an automated and adaptive AsyncOffloading strategy. We successfully solved the large model deployment problem on memory-limited devices with a performance drop of less than 2\%.

\textbf{Hybrid Parallelism for Distributed VAE.} Moreover, to address the issue of high GPU memory consumption due to the VAE-Decoder, we implemented an adaptive hybrid parallel strategy. This method partitions the input data along the spatial and temporal dimensions simultaneously and employs efficient collective communication for Conv3D computation. Thus, we further improved parallel scaling performance.

\textbf{Pipeline Optimizations.} We adopted kernel fusion, quantization, parallelization, continuous batching, prefix caching, and other common techniques to improve the overall throughput of the prompt engineering effectively. Furthermore, to tackle the issue of low encoding efficiency in long videos, we have implemented video encoding acceleration solutions. 

These innovations have effectively boosted the E2E efficiency of the whole inference pipeline.

\section{Training Infrastructure}
\subsection{Pre-Training Optimization}
To support efficient large-scale pre-training of long-context video models on thousands of GPUs, we have designed a highly optimized training infrastructure.
Our system focuses on maximizing hardware efficiency, scalability, and robustness. It integrates high-performance kernel fusion, a hybrid parallelism strategy, multi-level activation checkpointing (MLAC), runtime-aware workload balancing, and multi-level fault tolerance. These components work together to ensure stable, high-throughput training under diverse workloads and hardware scales.

\textbf{High-Performance Kernel.}
To fully utilize GPU hardware resources, we combined \texttt{torch.compile} with handcrafted CUDA kernels for performance-critical operators. We identified memory-bound operations and fuse them into single CUDA kernels to minimize redundant memory access, such as rotary position encoding (RoPE) and normalization. These fused kernels store intermediate results in registers or shared memory, significantly improving arithmetic intensity and reducing global memory traffic by over 90\%.

\textbf{Parallelism Strategy.}
We adopted a hybrid parallelism strategy combining data parallelism and sequence parallelism to efficiently train long-context models on thousands of GPUs. Specifically, we employed Hybrid Sharded Data Parallelism (HSDP)~\citep{zhao2023pytorch} for memory-efficient weight sharding and mitigating performance degradation observed when scaling to over thousands of GPUs. 
For sequence parallelism, we followed the Ulysses~\citep{jacobs2023deepspeed} approach, sharding tokens across GPUs along the sequence and head dimensions to enable parallel processing of long video samples. 

\textbf{Multi-Levels Activation Checkpointing.}
Multi-Level Activation Checkpointing (MLAC)~\citep{seawead2025seaweed} policy is employed to reduce GPU memory pressure under negligible recomputation overhead during backpropagation. MLAC implements optimized asynchronous caching and prefetching mechanisms to maximize the overlap between memory transfers and forward/backward computation. We leveraged MLAC to prioritize offloading output tensors of the operators (ops) with the highest recomputation cost during model training, e.g., attention and FC2 layer in MLP module. Furthermore, MLAC was applied to offload input tensors of the activation checkpointing module to attain zero activation occupancy in GPU memory, which allows us to lower the degree of sequence parallelism and thereby reduce communication overhead.

\textbf{Workload Balance.}
Large-scale video pre-training often involves heterogeneous data types (e.g., long vs. short videos, varying resolution), which introduces significant computational imbalance across GPUs. To address this, we applied a runtime-aware workload balancing strategy~\citep{seawead2025seaweed}, leveraging an additional all-to-all communication step to distribute workload evenly across GPUs.  
This balancing strategy is performed within each batch to maintain data consistency, and is asynchronously precomputed in the background to avoid stalling the main training loop. 
Our approach significantly reduced inter-GPU idle time and improves overall training throughput.

\textbf{Fault Tolerance.}
In large-scale training jobs running on thousands of GPUs over extended periods, transient failures are inevitable. To ensure robustness, we integrated fault tolerance at multiple levels.
First, we implemented periodic checkpointing of both model and optimizer states, with full support for FSDP-sharded weights. The states of the dataloader were also saved to ensure bitwise-exact resumption.
Second, we conducted thorough machine health checks before launching each job to eliminate potential stragglers and faulty nodes.
Third, we reduced model initialization overhead to maximize effective training time. For example, we utilized PyTorch’s meta tensor initialization to directly load model parameters, eliminating the time typically spent on standard initialization.
Combined, these strategies enhanced training reliability and minimize the impact of hardware or software failures during prolonged distributed runs.

\subsection{Post-Training Optimization}
Post-training primarily consists of three phases: supervised fine-tuning, reinforcement learning, and distillation. During this stage, it is essential not only to optimize training efficiency but also to minimize GPU memory consumption (e.g., reducing peak memory usage and fragmentation) and enhance overall memory utilization. The suboptimal GPU memory utilization observed in the post-training stage primarily stems from three factors:

\begin{itemize}
    \item \textbf{Memory Contention}. During the reinforcement learning and distillation phases, GPU memory is sequentially and dynamically shared among various components, including the Text Encoder, DiT, VAE, reward models, and their corresponding activation tensors.
    \item \textbf{Complex Training Modes}. The coexistence of trainable and frozen model components complicates memory management and introduces additional optimization challenges.
    \item \textbf{Diverse Workloads}. The concurrent presence of both long and short video sequences creates variable memory demands, making traditional static memory optimization methods ineffective.
\end{itemize}

To effectively address these challenges, we have developed a dynamic memory management framework that incorporates CPU offloading and recomputation techniques. Additionally, we adopted the parallelization strategies previously used during pre-training, leveraging FSDP and sequence parallelism to enable efficient multi-node scaling.

\begin{itemize}
    \item \textbf{Memory Optimization}. To ensure simplicity and ease of use, we utilized PyTorch hooks to implement CPU offloading,  thereby minimizing intrusive modifications to user code. Through detailed profiling and modeling, we identified optimal CPU offloading and recomputation strategies. In addition, we applied localized static memory planning to mitigate memory fragmentation caused by frequent allocation and free of tensors with varying sizes. 
    \item \textbf{Parallelism Strategy}. To maximize hardware utilization, we configured different degrees of sequence parallelism across different models based on their computational characteristics. Additionally, we set \texttt{TORCH\_NCCL\_AVOID\_RECORD\_STREAMS=1} to eliminate delayed memory release issues. Additionally, we manually managed the \texttt{free\_event\_queue} to address the problem of delayed parameter release in FSDP when parameters are frozen. Furthermore, we utilized \texttt{register\_post\_backward\_reshard\_only\_hook} to adjust the order of memory allocation and release during backward computation under the frozen mode.
\end{itemize}

These optimizations ensure stable and efficient post-training performance, even in complex scenarios involving multiple model components and diverse video workloads.

\section{Model Performance}

This section provides a comprehensive evaluation of Seedance 1.0, structured as follows. In Section 7.1, we first present results from an external public evaluation platform, where Seedance 1.0 tops the leaderboards in both text-to-video and image-to-video. Section 7.2 details the internal evaluation, covering benchmark design, absolute scoring, and comparative analysis using the Good-Same-Bad (GSB) metric. The subsequent subsections highlight Seedance 1.0's strengths in multi-shot transitions and multi-style generation. The overall results are presented in Figure 1.

\subsection{Artificial Analysis Arena}

Artificial Analysis~\cite{aa_elo} has emerged as a widely recognized and trusted benchmarking platform, particularly in the domains of image and video generation. It offers an open arena in which various generative models are evaluated and scored by the public. Leveraging a large corpus of comparison results, the platform calculates Elo scores to reflect user preferences across different models. 
The Artificial Analysis Video Arena Leaderboard comprises two distinct tracks: text-to-video and image-to-video. Seedance 1.0 has participated in both categories. Some notable external competitors include Veo 3, Kling 2.0, Runway Gen4, OpenAI Sora, and Wan 2.1.

\begin{figure}[t]
    \centering
    \includegraphics[width=\linewidth]{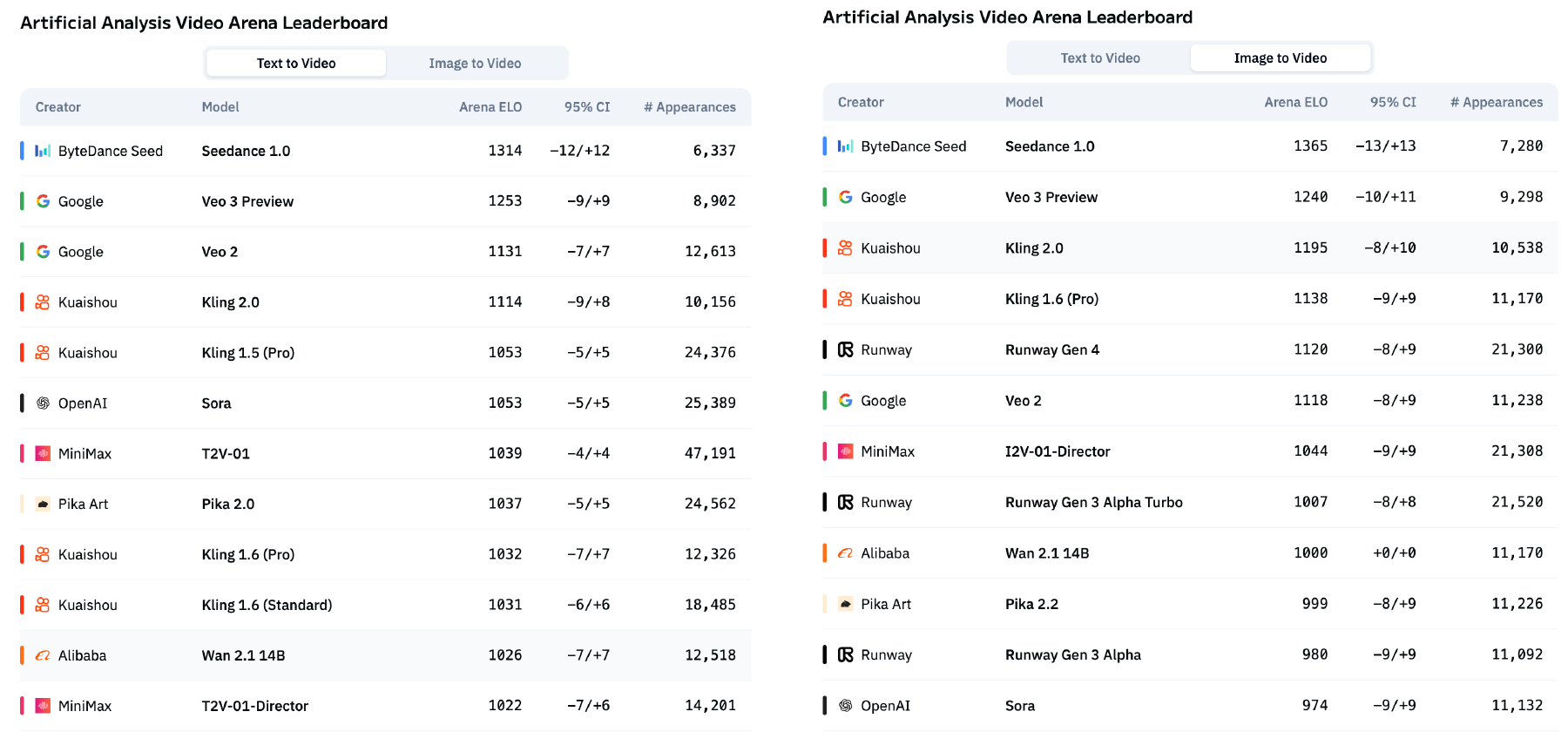}
    \caption{Results from Artificial Analysis Arena. Seedance 1.0 achieves the top position on both the text-to-video and image-to-video leaderboards.}
    \label{fig:aa_seedance}
\end{figure}

Seedance 1.0 tops both the text-to-video and image-to-video leaderboards, demonstrating a substantial performance advantage over competing models. In particular, it outperforms the second- and third-best models, Veo 3 and Kling 2.0, by over 100 points in the image-to-video task. Notably, Seedance 1.0 attains state-of-the-art results across both tasks using a single unified model, whereas prior models typically excelled in one domain while underperforming in the other. The subsequent sections provide a detailed analysis of Seedance 1.0's advantages in each scenario.

\subsection{Comprehensive Evaluation}
Besides overall user preferences, a comprehensive benchmark is equally important for the evaluation of visual generation models, as it enables a more holistic assessment of model capabilities. We developed SeedVideoBench-1.0, a comprehensive benchmark for video generation, comprising 300 prompts each for T2V and I2V. We then collaborated with film director experts to co-develop evaluation criteria and conducted a detailed manual expert evaluation.

\begin{figure*}[t]
\centering
\includegraphics[width=\linewidth]{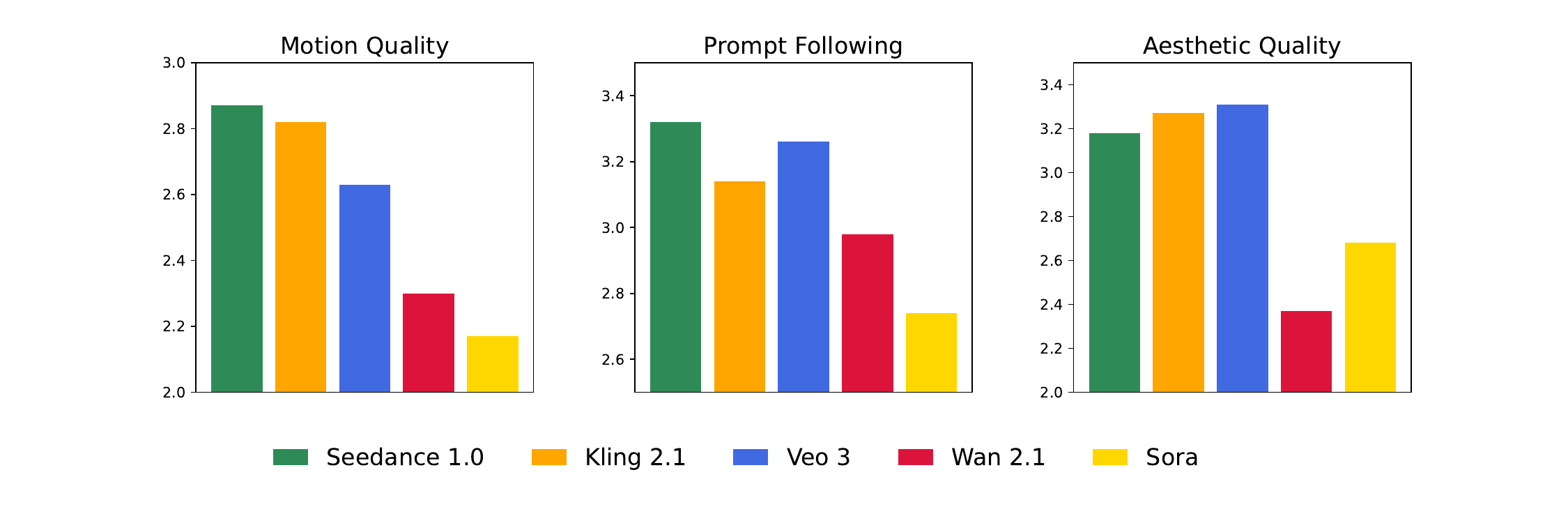}
\caption{Absolute Evaluation for Text-to-Video task.}
\label{fig:t2v_abs}
\end{figure*}

\begin{figure*}[!h]
\centering
\includegraphics[width=\linewidth]{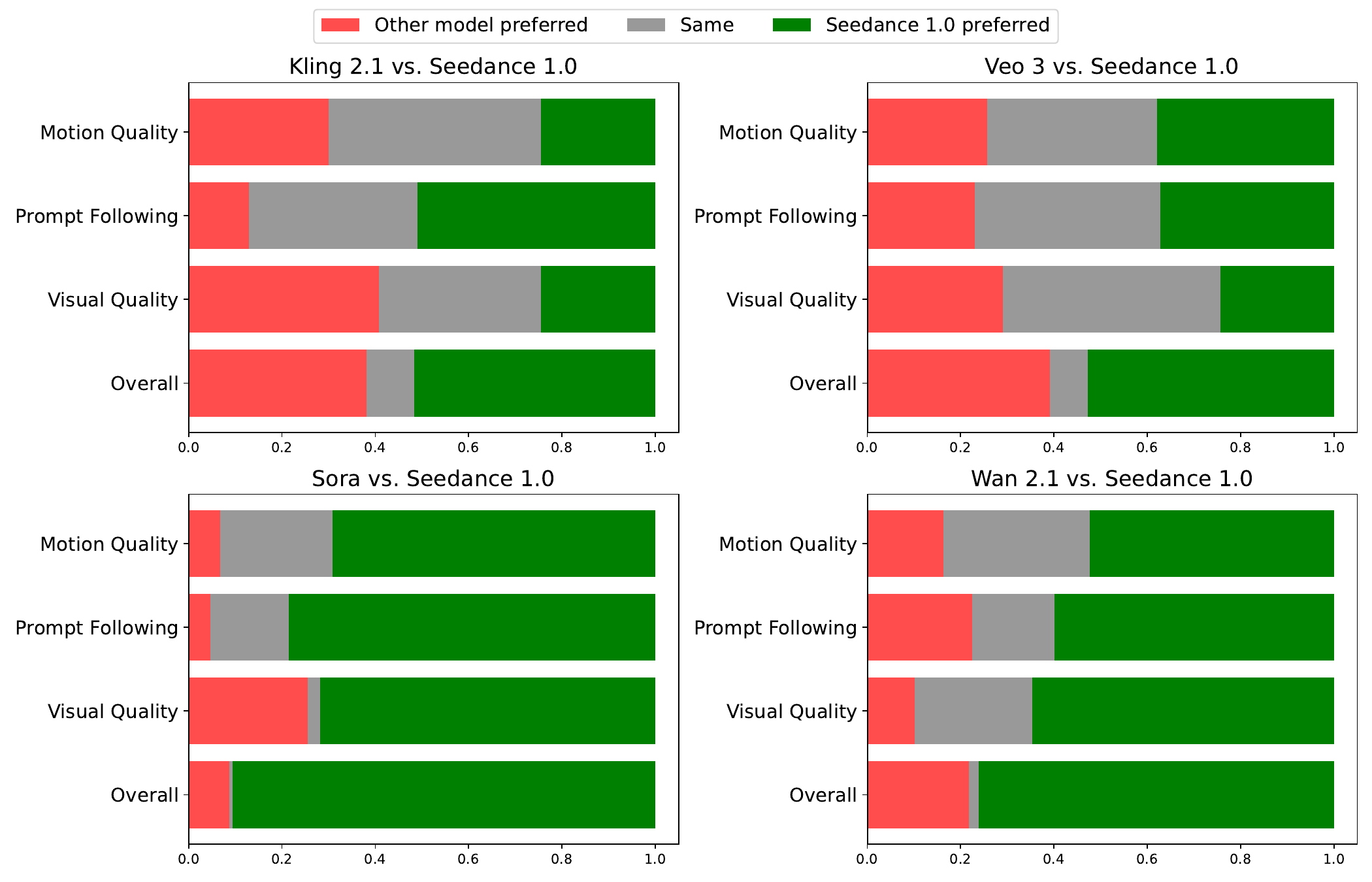}
\caption{GSB Evaluation for Text-to-Video Task.}
\label{fig:t2v_gsb}
\end{figure*}

\subsubsection{SeedVideoBench 1.0}
To comprehensively evaluate video generation models across diverse scenarios, we proposed SeedVideoBench-1.0, a benchmark designed through systematic analysis of real-world user prompts. This benchmark encompasses a wide range of application scenarios, including special effects, e-commerce, and professional-generated content (PGC). Additionally, a detailed taxonomy has been developed to assess model capabilities. The following section demonstrates the classification of main label categories, using text-to-video as an example.

\begin{itemize}[leftmargin=*]
\item ~\underline{\textit{Subject}} It is essential to first evaluate the model's ability to accurately generate primary entities, including humans, animals, natural scenes, consumer goods, and some virtual subjects.

\item ~\underline{\textit{Subject Description}} The focus is on models' ability to produce accurate representations of primary subjects. It includes subject quantity, entity attributes (e.g. appearance characteristics of human subjects, object properties of physical items), and spatial positioning.

\item ~\underline{\textit{Action}} Action simulation and generation represent fundamental capabilities of video generation models, indicative of their proficiency in capturing real-world dynamics and underlying physical laws. This category assesses motion-related actions across multiple categories, including human activities, multi-entity interactions, animal locomotion, sports movements, natural phenomena (e.g., weather events, biological processes), physical principles (e.g., gravity, fluid dynamics), and creative or imaginative motion patterns.

\item ~\underline{\textit{Action Description}} This category provides a finer-grained analysis of action generation, focusing on action number, movement direction, temporal sequencing, motion intensity, and expression of emotional states.

\item ~\underline{\textit{Camera}} The camera language component reflects a distinctive dimension of artistic expression in video generation, encompassing camera movements, shooting angles, shot size definition and variation, as well as transitions between multiple shots. SeedVideoBench-1.0 integrates a range of professional camera movements, including circular tracking shots, dolly-in shots, Hitchcock zooms, lateral pans, and follow shots.

\item ~\underline{\textit{Aesthetic description}} Aesthetics evaluation is an essential component in assessing visual generation models. This part encompasses style consistency, compositional atmosphere, lighting and shadow dynamics, and other factors governing the overall aesthetic quality of the generated videos.

\end{itemize}

The taxonomy for image-to-video is similar, with the addition of a labeling system for the first frame. For both text-to-video and image-to-video tasks, we construct 300 prompts each, uniformly distributed across the aforementioned categories. The quantity of prompts per category is designed to ensure sufficient discriminative and statistical confidence in the evaluation.

\subsubsection{Video Evaluation Metrics}
In collaboration with film directors, we developed a set of specialized evaluation metrics for generated videos, enabling assessment from a professional perspective. Unlike public preference evaluations, which often emphasize aesthetic appeal while neglecting fine-grained distinctions in model capabilities, this framework is structured around four core dimensions.

\begin{itemize}[leftmargin=*]

\item ~\underline{\textit{Motion Quality}} Motion Quality is the first intuitive impression that generated videos bring to users. It includes multiple aspects such as structural accuracy, motion plausibility, motion stability, and motion vividness. Structural accuracy focuses on detecting structural anomalies in generated content, such as extra limbs, truncation, unnatural bending, or inhuman postures. Motion plausibility involves physical plausibility in trajectory and speed, adherence to physical laws and common sense, and the identification of unnaturally static subjects or those with insufficient movement amplitude. Separately, motion stability is evaluated to detect artifacts caused by subject or background dynamics, while motion vividness addresses the coherence and realism of action sequences, including macro-structural integrity and the aesthetic quality of camera motion.

\item ~\underline{\textit{Prompt Following}} Prompt Following represents a foundational capability of generative models, reflecting their ability to produce content aligned with human intent. This evaluation focuses on multiple dimensions, including action responsiveness, subject description fidelity, stylistic conformity, incorporation of auxiliary entities, temporal alignment of motion, camera behavior, and environmental depiction accuracy.

\item ~\underline{\textit{Aesthetic Quality}} Evaluation of aesthetic appeal and visual quality in generated video emphasizes visual texture, perceptibility of AI sense, material detail fidelity, and the artistic expression of aesthetic intent.

\item ~\underline{\textit{Preservation}} Original image preservation, specific to image-to-video tasks, is assessed across multiple dimensions, including subject consistency, stylistic coherence, material fidelity, visual content alignment, and consistency in color and lighting.

\end{itemize}

\begin{figure*}[t]
\centering
\includegraphics[width=\linewidth]{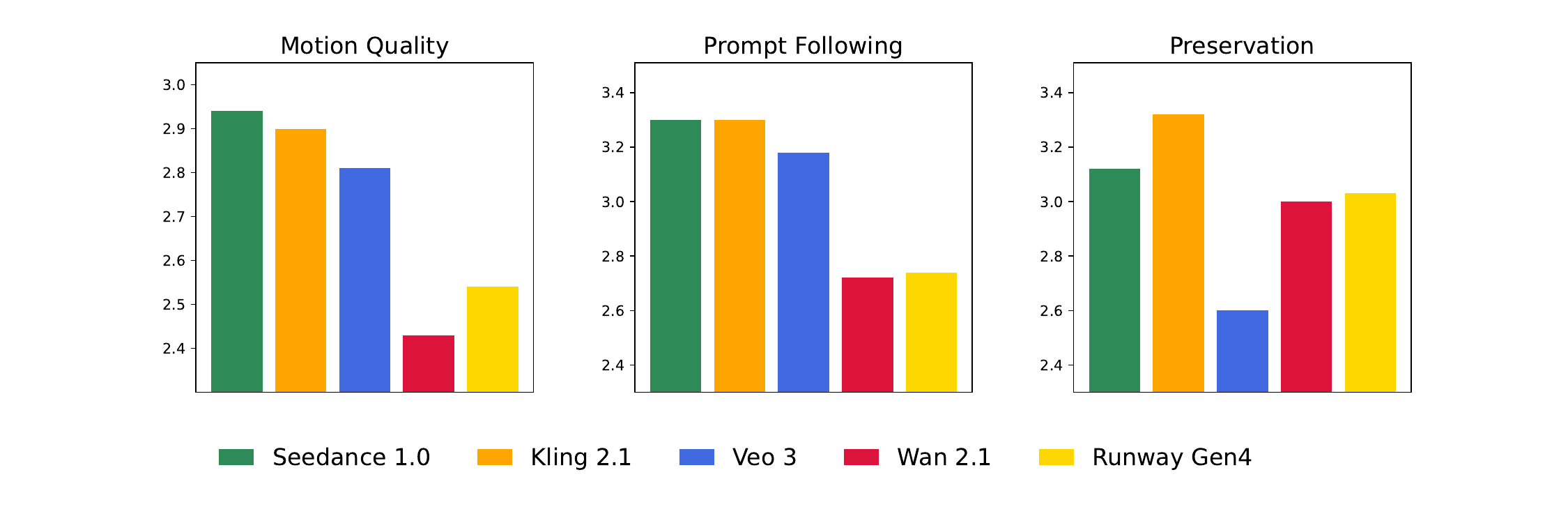}
\vspace{-1cm}
\caption{Absolute Evaluation for Image-to-Video task.}
\label{fig:i2v_abs}
\end{figure*}

\begin{figure*}[!h]
\centering
\includegraphics[width=\linewidth]{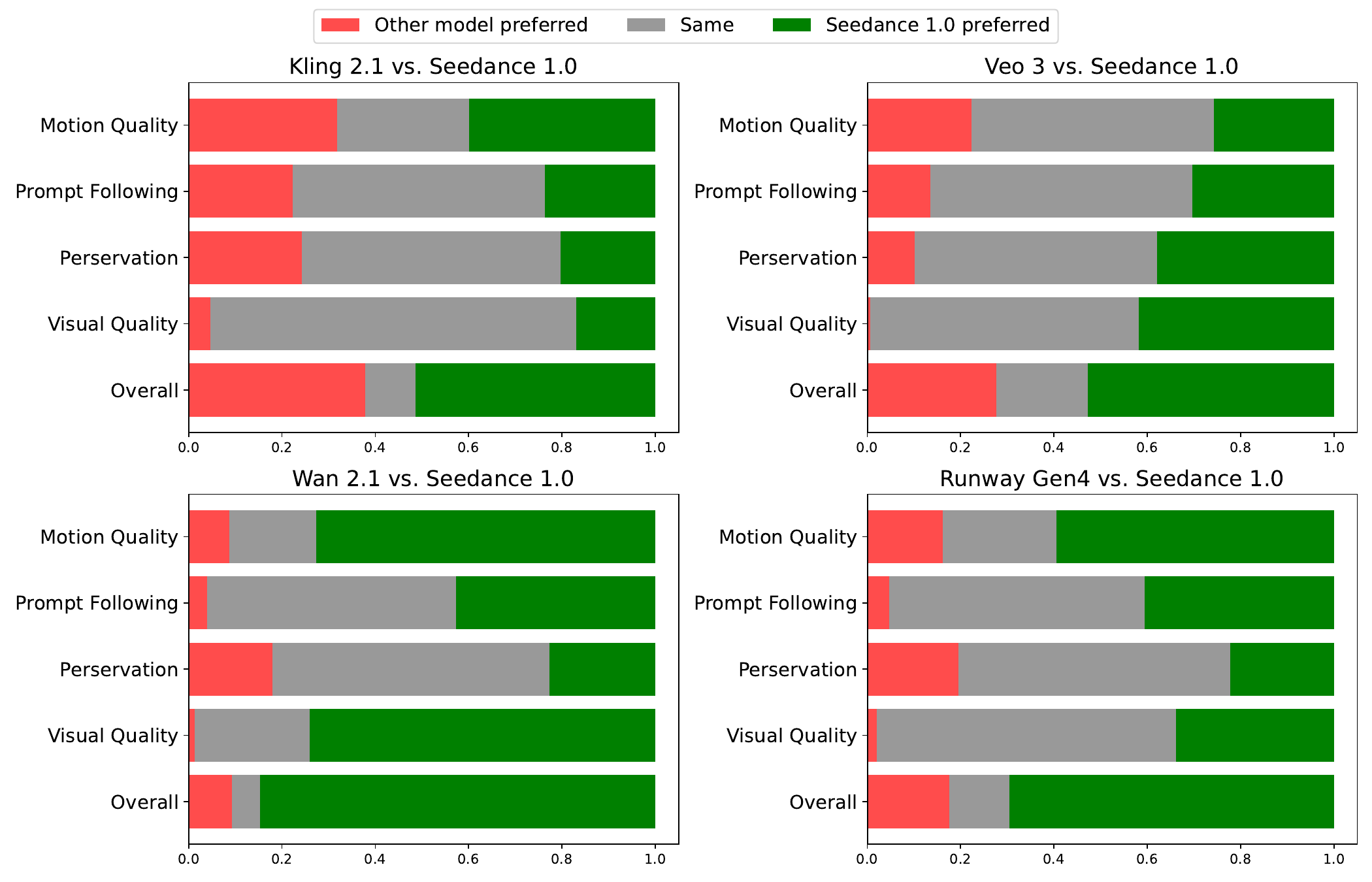}
\caption{GSB Evaluation for Image-to-Video task.}
\label{fig:i2v_gsb}
\end{figure*}

\subsubsection{Human Evaluation}

Leveraging SeedVideoBench 1.0, we conducted a comprehensive comparative evaluation of Seedance 1.0 against several leading video generation models across two tasks: text-to-video and image-to-video generation. For the text-to-video task, comparative models include Kling 2.1(Master), Veo 3, Wan 2.1, and Sora; for the image-to-video task, Sora is replaced by Runway Gen4. Two evaluation protocols are adopted: Absolute Score and the Good-Same-Bad (GSB) comparison metric. The Absolute Score employs a five-point Likert scale (where 1 indicates extreme dissatisfaction and 5 signifies utmost satisfaction), facilitating unified performance comparison across models. The GSB metric conducts pairwise comparisons to assess relative video quality, enabling fine-grained differentiation between model outputs.

\begin{figure*}[t]
\centering
\includegraphics[width=\linewidth]{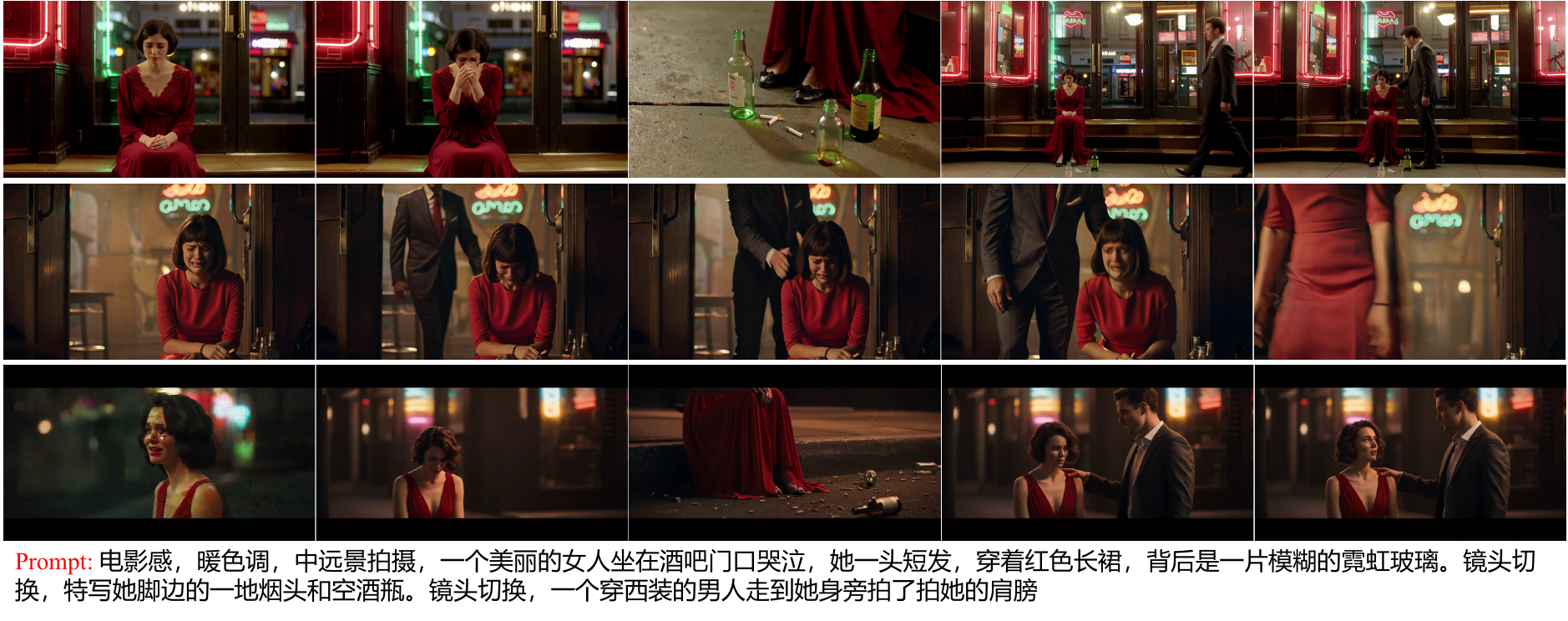}
\caption{Comparison of Multi-Shot Generation. Top: Seedance 1.0; Middle: Kling 2.1; Bottom: Veo 3.}
\label{fig:multi_shot_woman}
\end{figure*}

\begin{figure*}[!hb]
\centering
\includegraphics[width=\linewidth]{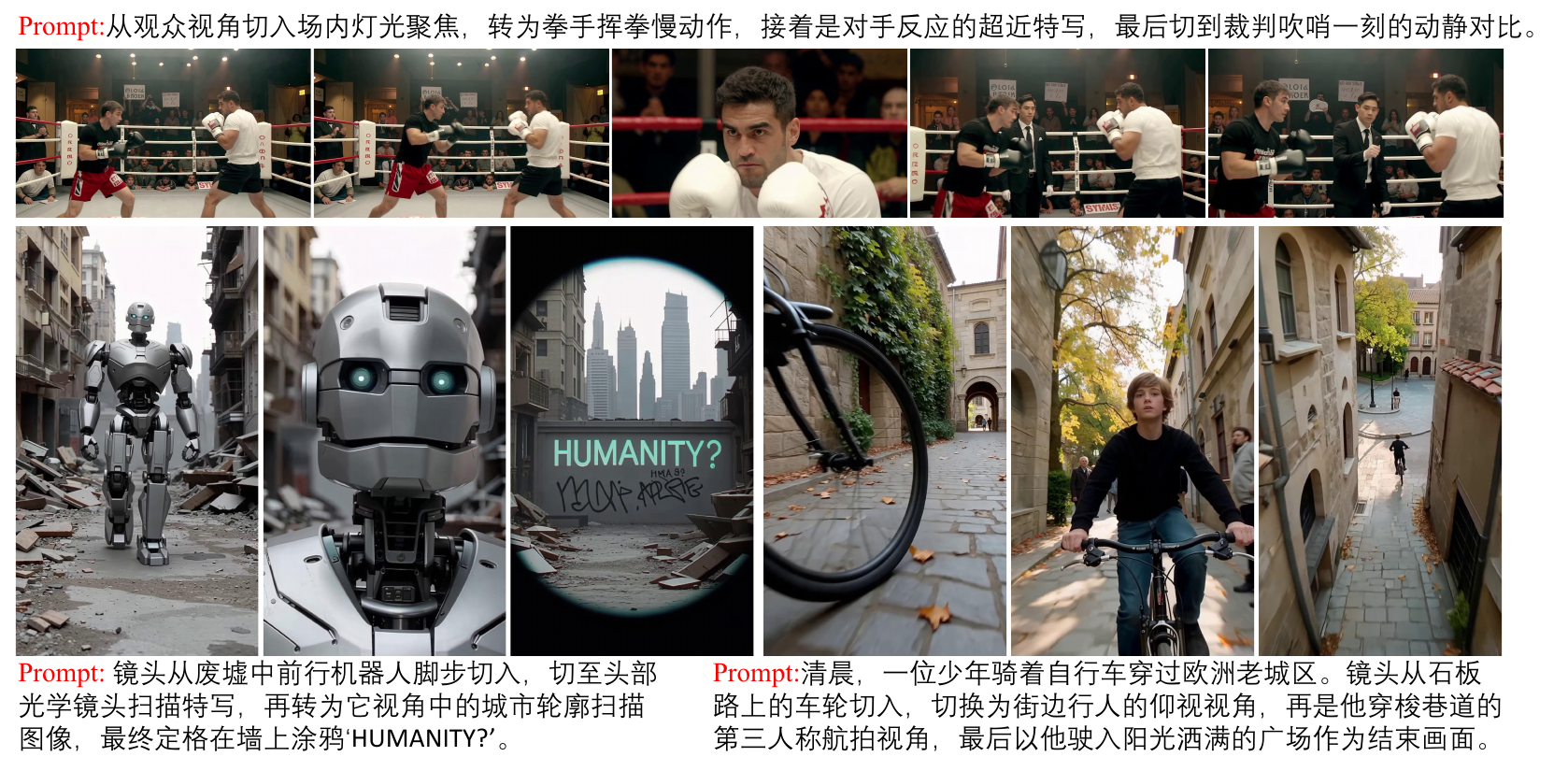}
\caption{Multi-Shot Generation for Seedance 1.0.}
\label{fig:multi_shot_showcase}
\end{figure*}

Figures \ref{fig:t2v_abs} and \ref{fig:t2v_gsb} show the absolute scores and GSB results for video generation models in the text-to-video task. Seedance 1.0, Kling 2.1, and Veo 3 substantially outperform other models. While Kling 2.1 demonstrates strong motion quality and visual fidelity, its limited prompt-following capability negatively impacts its overall effectiveness. In text-to-video generation, precise instruction adherence is critical to the adoption of generated content. Seedance 1.0 and Veo 3 exhibit superior prompt-following capability, driving their higher rankings on the Artificial Analysis leaderboard. Veo 3 excels at generating realistic videos, but its comparatively weaker motion quality constrains its capacity for complex video synthesis.

Figures \ref{fig:i2v_abs} and \ref{fig:i2v_gsb} present the absolute scores and GSB results for the image-to-video task. Seedance 1.0 and Kling 2.1 exhibit strong overall performance in this scenario. Adding image input as a condition introduces challenges in preserving character and background. Veo 3 performs relatively weak in this regard, occasionally altering lighting conditions, object textures, and other visual elements of the reference image. Additionally, it suffers from some quality degradation issues such as oily appearance or blurred details, which substantially affect its overall effectiveness. Kling 2.1 excels in motion quality, producing natural and coherent dynamics suitable for complex scenarios, though it occasionally experiences detail breakdown. Seedance 1.0 matches Kling 2.1's motion quality while offering superior prompt-following capability in scenarios involving complex shot transitions or detailed instruction prompts, resulting in more favorable overall performance.

\subsection{Multi-Shot Generation}
Seedance 1.0 demonstrates the capability to generate multiple consecutive shots from a single prompt, while ensuring subject continuity and stylistic coherence across frames. This enables the model to handle complex narrative techniques commonly used in cinematic storytelling. Specifically, Seedance 1.0 facilitates the construction of shot-reverse shot sequences for dialogic interaction, as well as the use of cut-in and cut-away shots to enrich narrative pacing and contextual layering. Furthermore, it supports match cuts and action cuts, enabling seamless transitions and preserving visual continuity. These competencies highlight Seedance’s proficiency in cinematic shot composition and temporal coherence, offering enhanced creative control and narrative expressiveness for video content generation. Figure~\ref{fig:multi_shot_woman} presents an example of continuous shot transitions generated by Seedance 1.0, which exhibits more coherent and fluid cinematic storytelling compared to other models.

\begin{figure*}[h]
\centering
\includegraphics[width=\linewidth]{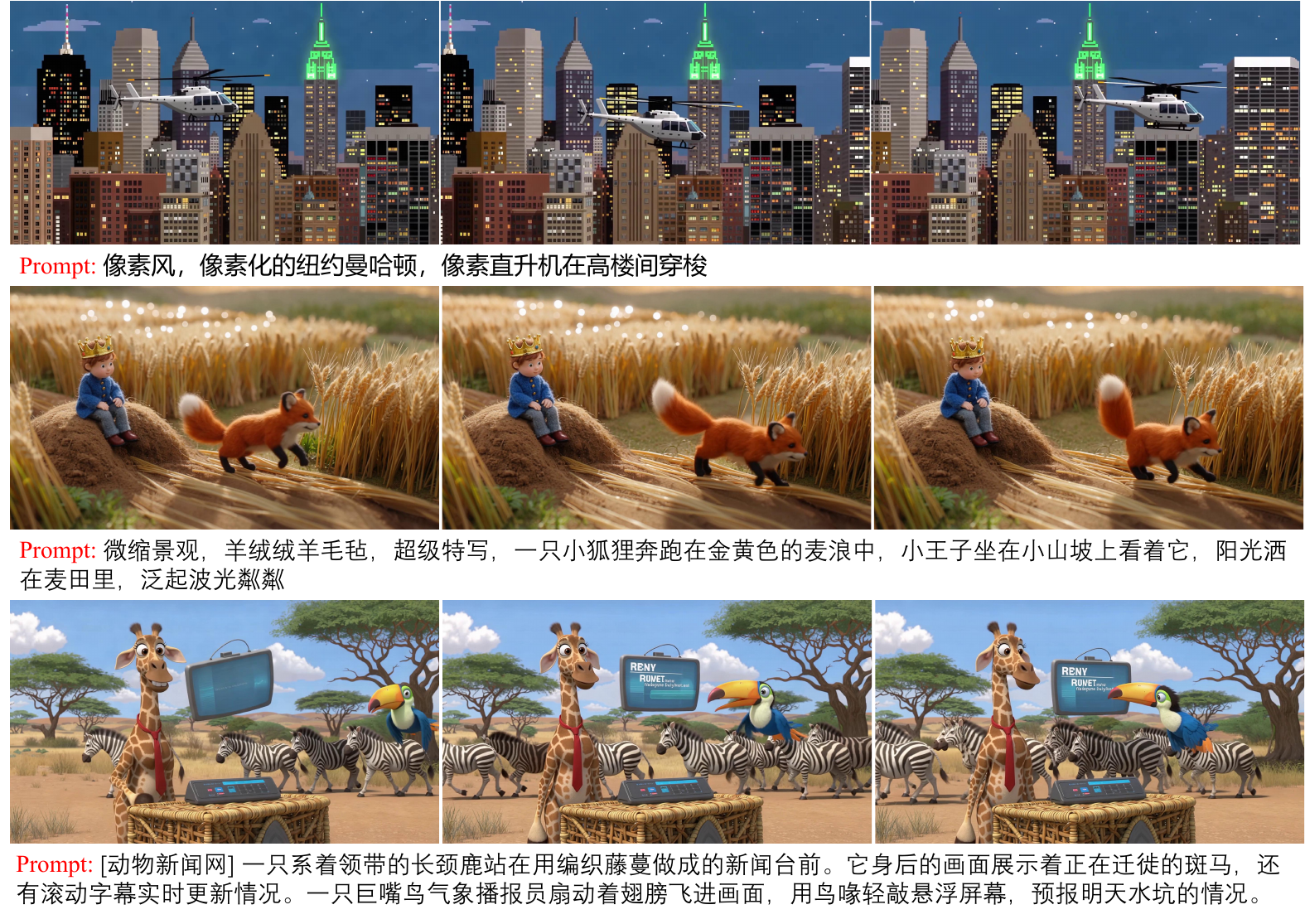}
\caption{Multi-Style Generation for Seedance 1.0.}
\label{fig:style_showcase}
\end{figure*}

\subsection{Multi-Style Alignment}
Seedance 1.0 exhibits strong generalization across a broad spectrum of visual styles. In text-to-video (T2V) tasks, Seedance 1.0 enables direct generation of fine-grained stylistic videos, while in image-to-video (I2V) tasks, it reliably preserves visual characteristics of the reference image. The model supports a wide range of real-world cinematic styles, including black-and-white silent films, classic Hong Kong cinema, and retro Hollywood aesthetics, as well as animated and fantasy-oriented styles such as Japanese anime, cyberpunk futurism, and ink-wash animation. This multi-style adaptability facilitates seamless transitions between realism and fantasy without the need for extensive task-specific tuning. As a result, Seedance 1.0 offers exceptional versatility and controllability, making it well-suited for professional filmmaking and AIGC creation.

\subsection{Visualization}
We present several visual outcomes by Seedance 1.0 in Figure \ref{fig:multi_shot_woman},\ref{fig:multi_shot_showcase},\ref{fig:style_showcase}.
For additional examples, please refer to the official website for an enhanced viewing experience.

\label{Conclusion}

\section{Conclusion} 

We have introduced Seedance 1.0, a native bilingual video generation foundation model that unifies multiple generation paradigms (such as text-to-video and image-to-video) and excels in instruction following, motion stability, and visual quality. We presented our technical improvements in dataset construction, efficient architecture design with training paradigm, post-training optimization, and inference acceleration, which are integrated effectively to achieve a high-performance model with fast inference. It demonstrates excellent capabilities in handling complex scenarios, multi-shot generation, and long-range temporal coherence, all while delivering fast and photorealistic generation experiences.

\clearpage

\bibliographystyle{plainnat}
\bibliography{main}

\clearpage

\beginappendix
\section{Contributions and Acknowledgments}
\label{contributions}

All contributors of Seedance are listed in alphabetical order by their last names.

\begin{multicols}{2} %
\sffamily{\color{seedblue}  \large{Core Contributors}} \\
\\
\color{seedblue}Yu Gao\\
\color{seedblue}Haoyuan Guo\\
\color{seedblue}Tuyen Hoang\\
\color{seedblue}Weilin Huang\\
\color{seedblue}Lu Jiang\\
\color{seedblue}Fangyuan Kong\\
\color{seedblue}Huixia Li\\
\color{seedblue}Jiashi Li\\
\color{seedblue}Liang Li\\
\color{seedblue}Xiaojie Li\\
\color{seedblue}Xunsong Li\\
\color{seedblue}Yifu Li\\
\color{seedblue}Shanchuan Lin\\
\color{seedblue}Zhijie Lin\\
\color{seedblue}Jiawei Liu\\
\color{seedblue}Shu Liu\\
\color{seedblue}Xiaonan Nie\\
\color{seedblue}Zhiwu Qing\\
\color{seedblue}Yuxi Ren\\
\color{seedblue}Li Sun\\
\color{seedblue}Zhi Tian\\
\color{seedblue}Rui Wang\\
\color{seedblue}Sen Wang\\
\color{seedblue}Guoqiang Wei\\
\color{seedblue}Guohong Wu\\
\color{seedblue}Jie Wu\\
\color{seedblue}Ruiqi Xia\\
\color{seedblue}Fei Xiao\\
\color{seedblue}Xuefeng Xiao\\
\color{seedblue}Jiangqiao Yan\\
\color{seedblue}Ceyuan Yang\\
\color{seedblue}Jianchao Yang\\
\color{seedblue}Runkai Yang\\
\color{seedblue}Tao Yang\\
\color{seedblue}Yihang Yang\\
\color{seedblue}Zilyu Ye\\
\color{seedblue}Xuejiao Zeng\\
\color{seedblue}Yan Zeng\\
\color{seedblue}Heng Zhang\\
\color{seedblue}Yang Zhao\\
\color{seedblue}Xiaozheng Zheng\\
\color{seedblue}Peihao Zhu\\
\color{seedblue}Jiaxin Zou\\
\color{seedblue}Feilong Zuo \\
\\

\noindent
\sffamily{\color{seedblue}  \large{Contributors}} \\
\\
\color{black}Sheng Bi\\
\color{black}Hao Chen\\
\color{black}Haoshen Chen\\
\color{black}Haoxin Chen\\
\color{black}Xiaoya Chen\\
\color{black}Feng Cheng\\
\color{black}Xuyan Chi\\
\color{black}Xiaojing Dong\\
\color{black}Junliang Fan\\
\color{black}Jing Fang\\
\color{black}Liangke Gui\\
\color{black}Qiushan Guo\\
\color{black}Bibo He\\
\color{black}Ruoqing Hu\\
\color{black}Siqi Jiang\\
\color{black}Ashley Kim\\
\color{black}Gen Li\\
\color{black}Yiying Li\\
\color{black}Haibin Lin\\
\color{black}Feng Ling\\
\color{black}Gaohong Liu\\
\color{black}Zuxi Liu\\
\color{black}Zhibei Ma\\
\color{black}Yanghua Peng\\
\color{black}Lei Shi\\
\color{black}Zuquan Song\\
\color{black}Renfei Sun\\
\color{black}Qinlong Wang\\
\color{black}Xuanda Wang\\
\color{black}Xun Wang\\
\color{black}Ye Wang\\
\color{black}Meng Wei\\
\color{black}Yawei Wen\\
\color{black}Ruolan Wu\\
\color{black}Xiaohu Wu\\
\color{black}Yonghui Wu\\
\color{black}Xin Xia\\
\color{black}Tingshuai Yan\\
\color{black}Zhouqike Yang\\
\color{black}Ziyan Yang\\
\color{black}Linxiao Yuan\\
\color{black}Zhonghua Zhai\\
\color{black}Manlin Zhang\\
\color{black}Xinyan Zhang\\
\color{black}Xinyu Zhang\\
\color{black}Zixiang Zhang\\
\color{black}Qi Zhao\\
\color{black}Rui Zhu\\
\color{black}Wenjia Zhu
\end{multicols} %

\end{CJK*}
\end{document}